\documentclass[11pt]{article}

\usepackage[final]{acl}

\usepackage{times}
\usepackage{latexsym}

\usepackage[T1]{fontenc}
\usepackage[utf8]{inputenc}

\usepackage{microtype}

\usepackage{inconsolata}

\usepackage{graphicx}

\usepackage{multirow}
\usepackage{booktabs}
\usepackage{threeparttable}
\usepackage{enumitem}
\usepackage{amsmath}
\usepackage{amssymb}

\usepackage{soul}
\usepackage{xcolor}
\title{Layer-Order Inversion: \\Rethinking Latent Multi-Hop Reasoning in Large Language Models}

\author{
Xukai Liu$^{1,\ast}$, Ye Liu$^{1,\ast}$, 
Jipeng Zhang$^{2}$, Yanghai Zhang$^{1}$, Kai Zhang$^{1}$, 
Qi Liu$^{1,\dagger}$ \\
$^{1}$State Key Laboratory of Cognitive Intelligence, 
University of Science and Technology of China \\
$^{2}$The Hong Kong University of Science and Technology \\
\texttt{\{chthollylxk,yhzhang0612\}@mail.ustc.edu.cn}, 
\texttt{yeliu.liuyeah@gmail.com}\\
\texttt{jzhanggr@connect.ust.hk},
\texttt{\{kkzhang08,qiliuql\}@ustc.edu.cn}
}

\begin{document}
\maketitle
\begin{abstract}
Large language models (LLMs) perform well on multi-hop reasoning, yet how they internally compose multiple facts remains unclear. Recent work proposes \emph{hop-aligned circuit hypothesis}, suggesting that bridge entities are computed sequentially across layers before later-hop answers. Through systematic analyses on real-world multi-hop queries, we show that this hop-aligned assumption does not generalize: later-hop answer entities can become decodable earlier than bridge entities, a phenomenon we call \emph{layer-order inversion}, which strengthens with total hops. To explain this behavior, we propose a \emph{probabilistic recall-and-extract} framework that models multi-hop reasoning as broad probabilistic recall in shallow MLP layers followed by selective extraction in deeper attention layers. 
Through probing and causal intervention analyses, we further show that shallow recall of deeper-hop knowledge can act as an intuition-like signal for answer selection, with stronger effects in weaker models and higher-hop queries.
Our framework further explains prior empirical phenomena and provides a mechanistic diagnosis of multi-hop failures despite correct single-hop knowledge.
% Code is available at \url{https://anonymous.4open.science/r/Layer-Order-Inversion/}.
Code is available at \url{https://github.com/laquabe/Layer-Order-Inversion}.
\end{abstract}

\renewcommand{\thefootnote}{\fnsymbol{footnote}}
\footnotetext[1]{Equal contribution.}
\footnotetext[2]{Corresponding author.}
\renewcommand{\thefootnote}{\arabic{footnote}}

\section{Introduction}
Large language models (LLMs) have demonstrated strong multi-hop reasoning capabilities~\cite{plaat2025multi-reasoning-survey1,liuxukai2025know3,liuxukai-etal-2024-onenet}, where answering a question requires integrating multiple pieces of factual knowledge. Such reasoning underlies many knowledge-intensive applications~\cite{liuyeah2025detect,liuyeah2023enhancing,liuxukai-etal-2023-rhgn,zhangyanghai-etal-2024-leveraging}.
Despite strong performance, how LLMs internally perform multi-hop reasoning remains poorly understood~\cite{press2023Measuring-and-narrowing-the-compositionality-gap-in-language-models, hou2023towards-Interpretation-Multi-Step-Reasoning}.

% Large language models (LLMs) have demonstrated strong multi-hop reasoning capabilities~\cite{plaat2025multi-reasoning-survey1,wu2025mmqa,li-etal-2024-making}, where answering a question requires integrating multiple pieces of factual knowledge. Such reasoning underlies many knowledge-intensive applications~\cite{wei-etal-2025-plangenllms,guo2024llm-multi-agents-survey,yu2024evaluation-rag-survey2}.
% Despite strong performance, how LLMs internally perform multi-hop reasoning remains poorly understood~\cite{press2023Measuring-and-narrowing-the-compositionality-gap-in-language-models, hou2023towards-Interpretation-Multi-Step-Reasoning}.

\begin{figure}[t]
    \centering
    \includegraphics[width=1\linewidth]{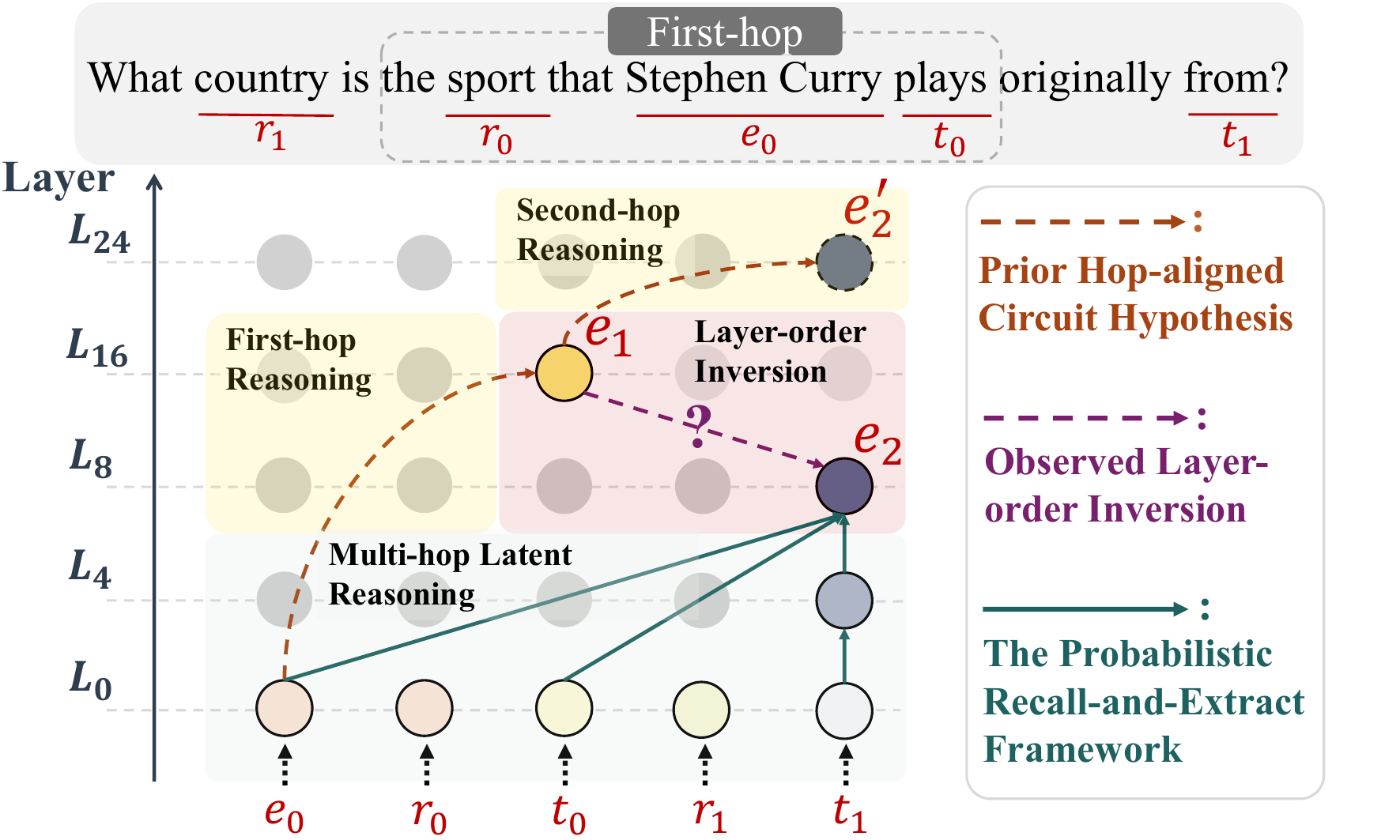}
    \caption{An illustration of latent multi-hop reasoning.
The \textit{layer-order inversion} contradicts hop-aligned circuits, while our \textit{probabilistic recall-and-extract} framework interprets it as horizontal recall of deeper-hop knowledge that may act as an intuition-like signal.
    }
    \label{fig:intro}
\end{figure}

Recent work has proposed that LLMs realize multi-hop reasoning through explicit \emph{reasoning circuits}~\cite{wang2024grokking, yao2024knowledge-circuits, yao-etal-2025-cake}, in which each hop is resolved sequentially across layers.
Under this view, as illustrated in Figure~\ref{fig:intro}, bridge entities (e.g., $e_1$) are computed at earlier layers and propagated forward, yielding a hop-aligned, layer-wise computation ($e_1 \rightarrow e_2'$).
Empirical support for this hypothesis largely comes from analyses on two-hop or synthetic datasets, using techniques such as layer-wise decoding~\cite{ghandeharioun2024patchscopes, belinkov2022probing}.

However, whether LLMs’ latent reasoning follows such hop-aligned circuits remains unclear beyond restricted settings~\cite{biran2024hopping-too-late, yang-2024-Do-Large-Language-Models-Latently-Perform-Multi-Hop-Reasoning}. As illustrated in Figure~\ref{fig:intro}, the reasoning-circuit view predicts a strict ordering in which bridge entities (e.g., \(e_1\)) emerge at earlier layers than later-hop entities (e.g., \(e_2'\)). In contrast, we observe clear counterexamples: the final answer entity (e.g., \(e_2\)) can become decodable at earlier layers than the corresponding bridge entity \(e_1\). We refer to this phenomenon as \emph{layer-order inversion}. 
Moreover, hop-aligned circuits also struggle to explain reasoning phenomena that are not primarily tied to model depth like shortcuts~\cite{yuan-etal-2024-llms-shortcut}.
These limitations motivate us to re-examine the hop-aligned circuit hypotheses.

To this end, we systematically investigate layer-order inversion on MQuAKE~\cite{zhong-etal-2023-mquake}, a benchmark with up to four-hop queries and annotated intermediate facts.
Using Patchscopes~\citep{ghandeharioun2024patchscopes}, we show that layer-order inversion becomes increasingly pronounced in higher-hop queries, providing evidence that hop-aligned circuits are insufficient as a general explanation of latent multi-hop reasoning.

To explain these findings, we extend previous hypotheses~\cite{geva2023dissecting-recall-then-extract} and propose a \emph{probabilistic recall-and-extract} framework that views latent multi-hop reasoning as probabilistic knowledge recall and extraction rather than deterministic hop-by-hop computation.
In this framework, knowledge is recalled not only vertically across layers but also horizontally across tokens. Supported by Patchscopes and hidden-state similarity analyses, horizontal recall explains layer-order inversion by allowing deeper-hop knowledge to become available before all bridge entities are explicitly decoded. Causal intervention further demonstrates that such shallow horizontal recall can act as an intuition-like signal for answer selection, while deeper attention layers selectively extract accumulated answer-relevant knowledge. This perspective offers a more complete account of prior observations: hop-aligned circuits can be interpreted as dominant recall patterns, whereas shortcut behaviors and chain-of-thought prompting reflect the role of horizontal recall and intuition-like signals in shaping answer selection. It also explains multi-hop failures as insufficient recall or extraction under the multi-hop context, even when all relevant single-hop facts are answered correctly.

Our contributions are summarized as follows:
\begin{itemize}[leftmargin=*, itemsep=0pt]
    \item For the first time, we uncover \emph{layer-order inversion}, where later-hop entities become decodable earlier than bridge entities in LLMs, demonstrating that hop-aligned circuits do not generalize to higher-hop queries.
    
    \item We introduce a \emph{probabilistic recall-and-extract} framework that models multi-hop reasoning as probabilistic recall and selective extraction, highlighting shallow recall of deeper-hop knowledge as an intuition-like mechanism.

    \item Through extensive experiments, we show that this framework effectively explains prior circuit-like observations, shortcuts, and systematic failure modes in multi-hop reasoning.
\end{itemize}

\section{Preliminaries}

\subsection{Problem Definition}

We consider multi-hop reasoning over a set of entities $\mathcal{E}$ and relations $\mathcal{R}$, where each single-hop fact is represented as a triplet $(e, r, e') \in \mathcal{E} \times \mathcal{R} \times \mathcal{E}$.
By treating each relation $r$ as a mapping function $r(e) = e'$, a $k$-hop fact can be defined as:
\begin{equation}
e_{i+1} = r_i(e_i), \quad  i = 0,\dots,k-1,
\end{equation}
which yields the chained expression:
\begin{equation}
    r_{k-1} \circ r_{k-2} \circ \cdots \circ r_0 (e_0) = e_{k},
\end{equation}
where $e_0$ is the subject entity, $e_k$ is the answer entity, and $\{e_1,\dots,e_{k-1}\}$ are the bridge entities.

While multi-hop reasoning can be formalized as relation composition, it remains unclear whether LLMs internally decompose a $k$-hop query into single-hop steps or implicitly encode the composed relation $r_{k-1} \circ \cdots \circ r_0$~\cite{dai2022knowledge-neurons}.
We therefore study latent multi-hop reasoning by tracing whether bridge entities ${e_1, \dots, e_{k-1}}$ emerge in hidden states, and where they appear~\cite{zhu2025survey-lantent-reasoning-survey1}.

% While multi-hop reasoning can be formalized as a composition of relations, it remains unclear whether LLMs internally decompose a $k$-hop query into a sequence of single-hop steps or instead encode the composed relation $r_{k-1} \circ \cdots \circ r_0$ implicitly~\cite{dai2022knowledge-neurons}. 
% Latent multi-hop reasoning therefore concerns tracing how an LLM internally processes a $k$-hop query~\cite{zhu2025survey-lantent-reasoning-survey1, chen2025reasoning-lantent-reasoning-survey2}. Specifically, we analyze whether the bridge entities ${e_1, \dots, e_{k-1}}$ emerge in the hidden states, and where they appear.

\begin{figure*}[!ht]
    \centering
    \includegraphics[width=\textwidth]{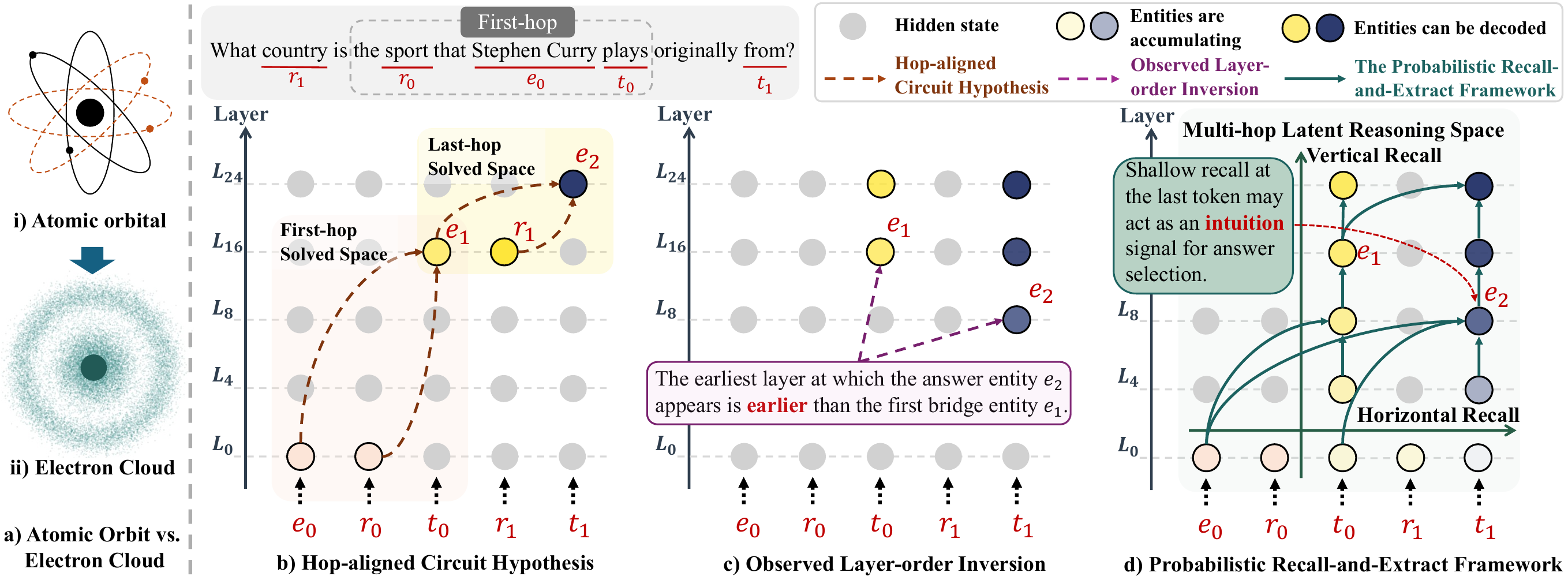}
    \caption{An illustration of latent multi-hop reasoning in LLMs from a probabilistic perspective. 
(a) Electron behavior follows a probabilistic distribution rather than fixed orbits. 
(b) \textbf{Hop-aligned circuit hypothesis} assumes layer-by-layer recall of bridge entities. 
(c) Observed \textbf{layer-order inversion}, where later-hop entities may emerge earlier than bridge entities. 
(d) \textbf{Probabilistic recall-and-extract framework} combining vertical and horizontal recall, with last token shallow recall serving as an intuition-like signal for answer selection.
}
    \label{fig:framework}
\end{figure*}

\subsection{Hop-aligned Circuits Hypothesis}
\label{Multi-hop Reasoning Circuits}
% Knowledge circuits refer to computation subgraphs in LLMs whose activations are sufficient to reproduce specific factual knowledge~\cite{yao2024knowledge-circuits}. Building on this notion, subsequent work has proposed \emph{hop-aligned circuits hypothesis}, which posit that multi-hop inference is realized by resolving each hop in a step-by-step manner across layers~\cite{wang2024grokking, yao-etal-2025-cake}.
% Formally, hop-aligned circuits assume that a $k$-hop query is resolved recursively as follows:

Knowledge circuits refer to computation subgraphs in LLMs whose activations are sufficient to reproduce specific factual knowledge~\cite{yao2024knowledge-circuits}. 
The \emph{hop-aligned circuits hypothesis} extends this view to multi-hop inference, positing that each hop is resolved sequentially across layers~\cite{wang2024grokking, yao-etal-2025-cake}. 
Formally, a $k$-hop query is resolved recursively:
\begin{equation}
    e_{i+1} = F_i(e_i, r_i), \quad i = 0,\dots,k-1,
\end{equation}
where $F_i$ denotes the mapping function from $(e_i, r_i)$ to $e_{i+1}$. The final answer is obtained as:
\begin{equation}
    F_{k-1}\bigl(F_{k-2}(\cdots F_0(e_0, r_0)\cdots, r_{k-1})\bigr)=e_k.
\end{equation}
Each $F_i$ is hypothesized to be implemented by a subset of layers, producing a circuit-like computation in which bridge entities are recalled and propagated in hop order, as illustrated in Figure~\ref{fig:framework}(b).

% Empirical support for hop-aligned circuits has been reported from multiple perspectives. Early studies~\cite{biran2024hopping-too-late, ghandeharioun2024patchscopes} show that patching hidden-state from single-hop queries or later layers into earlier layers can improve performance on multi-hop queries, suggesting that bridge entities are computed at shallow layers.
% Training-based studies~\cite{wang2024grokking, yao-etal-2025-language-models-multi-hop-reasoning-training-data} fine-tune models on synthetic multi-hop datasets and use causal tracing techniques~\cite{meng2022locating-rome} to reveal that positions with significant causal influence on the final prediction exhibit a layer-wise, hop-aligned distribution.
% In parallel, non-tuning approaches~\cite{yao-etal-2025-cake, zhanglocate-then-edit-for-Multi-hop-Factual-Recall} based on logit lens and patching techniques~\cite{geva-etal-2022-transformer-logit-len, wang2025logitlens4llms} report that bridge entities emerge at earlier layers than answer entities, a pattern interpreted as evidence for hop-aligned circuits.

Empirical support for hop-aligned circuits has been reported from multiple perspectives. 
Early studies~\cite{biran2024hopping-too-late, ghandeharioun2024patchscopes} show that patching hidden states from single-hop queries or later layers into earlier layers improves multi-hop performance, suggesting that bridge entities are computed in shallow layers.
Training-based studies~\cite{wang2024grokking, yao-etal-2025-language-models-multi-hop-reasoning-training-data} fine-tune models on synthetic multi-hop data and use causal tracing~\cite{meng2022locating-rome} to show that causally important positions exhibit hop-aligned distributions.
In parallel, non-tuning approaches~\cite{yao-etal-2025-cake, zhanglocate-then-edit-for-Multi-hop-Factual-Recall} use logit-lens and patching techniques~\cite{geva-etal-2022-transformer-logit-len, wang2025logitlens4llms} to show that bridge entities emerge earlier than answer entities, which is interpreted as evidence for hop-aligned circuits.

% Despite these findings, whether LLMs truly rely on hop-aligned circuits remains unclear.
% First, training-based evidence depends on explicit fine-tuning with synthetic multi-hop data and may not reflect the intrinsic behavior of pretrained models. 
% Second, non-tuning analyses are largely restricted to two-hop settings and fixed templates, which may bias conclusions. 
% Finally, prior work~\cite{yao2024knowledge-circuits} shows that models can still correctly answer certain multi-hop questions even after first-hop knowledge is removed, challenging strictly sequential and symbolic circuit-based explanations.
% Related phenomena, such as shortcut in multi-hop questions~\cite{ju2024investigating-Multi-Hop-Factual-Shortcuts, zhang2024enhancing-multi-hop-reasoning-knowledge-erasure} and the effects of chain-of-thought prompting~\cite{wei2022chain-of-thought, renze2024benefits-ccot}, further complicate the interpretation of circuit-based reasoning.
% These limitations motivate us to re-examine the validity of circuits, particularly for higher-hop queries, to better understand the latent multi-hop reasoning in LLMs.

Despite these findings, whether LLMs truly rely on hop-aligned circuits remains unclear. 
First, training-based evidence depends on explicit fine-tuning with synthetic multi-hop data and may not reflect the intrinsic behavior of pretrained models. 
Second, non-tuning analyses are largely restricted to two-hop settings and fixed templates, which may bias conclusions. 
Finally, prior work~\cite{yao2024knowledge-circuits} shows that models can still answer certain multi-hop questions correctly even after first-hop knowledge is removed, challenging strictly sequential circuit-based explanations. 
Related phenomena, such as shortcuts~\cite{ju2024investigating-Multi-Hop-Factual-Shortcuts, zhang2024enhancing-multi-hop-reasoning-knowledge-erasure} and chain-of-thought~\cite{wei2022chain-of-thought, renze2024benefits-ccot}, further complicate circuit-based interpretations.
These limitations motivate us to re-examine hop-aligned circuits, particularly for higher-hop queries, to better understand latent multi-hop reasoning in LLMs.

\subsection{Rethinking Latent Multi-hop Reasoning}
To this end, we investigate how LLMs internally perform multi-hop reasoning through the following research questions:

\noindent\textbf{RQ1: Does latent multi-hop reasoning in LLMs follow hop-aligned circuits?}

This question examines whether LLMs exhibit latent multi-hop reasoning when answering multi-hop queries, and whether such reasoning conforms to the layer-wise structure assumed by hop-aligned circuit hypothesis.

\noindent\textbf{RQ2: If hop-aligned circuits do not fully explain LLM behavior, what internal mechanisms support multi-hop reasoning?}

We aim to characterize the internal mechanisms by which LLMs acquire and utilize the knowledge required for multi-hop reasoning, and assess how these mechanisms can account for prior empirical observations.

\noindent\textbf{RQ3: Why do LLMs fail on multi-hop questions?}

We further investigate the causes of failure in multi-hop queries, particularly in cases where LLMs correctly answer all relevant single-hop facts but still produce incorrect multi-hop predictions.

\paragraph{Overview.}
To address these research questions, 
Section~\ref{sec:setup} introduces our experimental setup for analyzing latent multi-hop reasoning.
Section~\ref{Do LLMs Rely on Explicit Multi-hop Reasoning Circuits?} examines whether LLMs follow hop-aligned circuits (RQ1).
Section~\ref{Sec: A Probabilistic Perspective on the Recall-and-Extract Framework} investigates the internal mechanisms of multi-hop reasoning from a probabilistic perspective, and identifies an intuition-like phenomenon in which deeper-hop knowledge can be directly recalled in LLMs (RQ2).
Section~\ref{Sec: Reinterpreting Errors in Multi-hop Reasoning} analyzes failure cases on multi-hop questions (RQ3).
Finally, Section~\ref{sec:conclusion} concludes this paper.

\begin{table}[t]
\centering
\resizebox{\linewidth}{!}{
\begin{tabular}{l l r r r r}
\toprule
\textbf{Model} & \textbf{Category} & \textbf{2-hop} & \textbf{3-hop} & \textbf{4-hop} & \textbf{Total} \\
\midrule
\multirow{3}{*}{GPT-J-6B}
 & Correct   & 233 & 133 & 42  & 408 \\
 & Incorrect & 298 & 366 & 178 & 842 \\
 & Missing   & 562 & 605 & 460 & 1,627 \\
\midrule
\multirow{3}{*}{Llama3-8B}
 & Correct   & 609 & 330 & 275 & 1,214 \\
 & Incorrect & 420 & 477 & 397 & 1,294 \\
 & Missing   & 301 & 451 & 168 & 920 \\
\bottomrule
\end{tabular}
}
\caption{
Dataset statistics of MQuAKE by hop count and outcome category (\textbf{Correct}, \textbf{Incorrect}, \textbf{Missing}).
}
\label{tab:data_distribution}
\end{table}

\section{Experimental Setup}
\label{sec:setup}
% This section describes the datasets and models used in our study, as well as the analysis protocols for probing latent multi-hop reasoning in LLMs.

\subsection{Datasets and Models}

Our experiments are based on MQuAKE~\cite{zhong-etal-2023-mquake}, which contains multi-hop questions with up to four hops, multiple natural-language verbalizations per query, and explicit annotations of intermediate facts. These properties allow us to probe the internal emergence of bridge entities beyond two-hop, template-based synthetic settings.

% To analyze model behavior under different outcome, we partition MQuAKE into three subsets based on model predictions.
% \textbf{Correct} consists of instances where both the multi-hop question and all single-hop questions are answered correctly.
% \textbf{Incorrect} includes instances where the model fails on the multi-hop question despite correctly answering all single-hop questions.
% \textbf{Missing} includes instances where the model fails on the multi-hop question and answers at least one associated single-hop question incorrectly.
% Dataset statistics are summarized in Table~\ref{tab:data_distribution}.

To analyze outcome-specific behavior, we partition MQuAKE by model predictions into three subsets: \textbf{Correct} includes cases where both multi-hop and all single-hop questions are answered correctly; \textbf{Incorrect} includes cases where only the multi-hop question fails; and \textbf{Missing} includes cases where the multi-hop question and at least one associated single-hop question fail.
Dataset statistics are summarized in Table~\ref{tab:data_distribution}.
We evaluate two representative decoder-only LLMs: GPT-J-6B~\cite{gpt-j} and Llama~3-8B~\cite{touvron2023llama}.
We use greedy decoding (temperature \(=0\)) to ensure deterministic and reproducible inference.

\begin{figure*}[!ht]
    \centering
    \includegraphics[width=\textwidth]{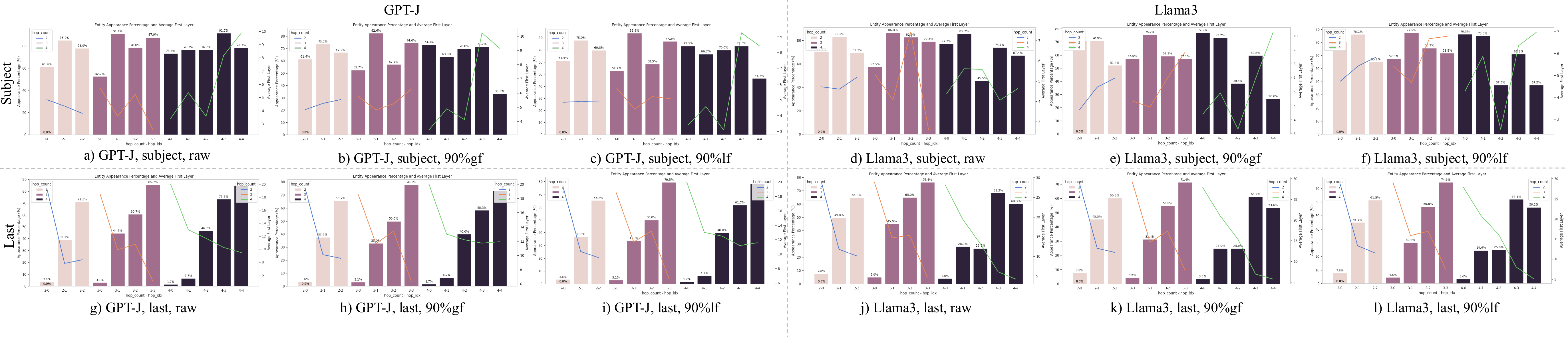}
    \caption{Patchscopes results for GPT-J (left) and Llama3 (right) on the Correct subset. We probe hidden states at the last subject token (top) and the last token of the query (bottom) under three settings: raw, 90\% gf, and 90\% lf.}
    \label{fig:patchscopes_class1}
\end{figure*}
\begin{table*}[t]
\centering
\small
\begin{tabular}{l l|rrr|rrrr|rrrrr}
\toprule
\multirow{2}{*}{Model} & \multirow{2}{*}{Token}
& \multicolumn{3}{c}{2-hop}
& \multicolumn{4}{c}{3-hop}
& \multicolumn{5}{c}{4-hop} \\
\cmidrule(lr){3-5} \cmidrule(lr){6-9} \cmidrule(lr){10-14}
& & $e_0$ & $e_1$ & $e_2$
  & $e_0$ & $e_1$ & $e_2$ & $e_3$
  & $e_0$ & $e_1$ & $e_2$ & $e_3$ & $e_4$ \\
\midrule
\multirow{2}{*}{GPT-J}
& Subject
& 4.83 & 4.35 & 3.81
& 5.74 & 3.60 & 5.24 & 2.55
& 3.41 & 5.37 & 3.59 & 8.18 & 9.89 \\
& Last
& 19.36 & 7.81 & \textbf{8.38}
& 18.57 & 9.91 & 10.76 & \textbf{5.05}
& 30.00 & 13.00 & 11.68 & 10.27 & \textbf{9.47} \\
\midrule
\multirow{2}{*}{Llama3}
& Subject
& 4.72 & 4.61 & 5.20
& 5.34 & 4.08 & 7.41 & 2.63
& 4.37 & 5.61 & 5.59 & 4.06 & 4.64 \\
& Last
& 28.92 & 11.81 & \textbf{10.11}
& 28.93 & 17.75 & 15.34 & \textbf{4.64}
& 28.22 & 19.43 & 13.18 & 6.08 & \textbf{4.24} \\
\bottomrule
\end{tabular}
\caption{Average Earliest Layer of Entity Emergence under the Raw Setting}
\label{tab:earliest_layer_raw}
\end{table*}

\subsection{Patchscopes Analysis}
\label{sec:patchscopes}

We adopt Patchscopes~\cite{ghandeharioun2024patchscopes} as our primary tool for probing latent multi-hop reasoning in LLMs. Given a selected token and layer, Patchscopes extracts the corresponding hidden state from the source query, patches it into a placeholder token in a target prompt, and uses the generation from the target prompt to probe the information encoded in the source hidden state. Compared to supervised probing methods~\cite{belinkov2022probing, belrose2023eliciting-tuned-lens}, Patchscopes is fully unsupervised and reveals when different entities become decodable across layers, especially in shallow layers.
Following prior work~\cite{biran2024hopping-too-late}, we use the same Patchscopes setup, repeating the patching procedure three times per query and aggregating the results across runs.

% \paragraph{Post-processing.}
% Patchscopes generations from shallow layers can be noisy due to insufficient contextual priors, which may spuriously overlap with entities of interest. To reduce such noise, we apply two filtering strategies.
% \begin{itemize}[leftmargin=*, itemsep=0pt]
%     \item \textbf{Global filtering (GF)} computes the similarity~\cite{reimers2019sentence-bert} between generated outputs and the original query, and removes the bottom $k\%$ of results with the lowest similarity across the entire dataset.
%     \item \textbf{Local filtering (LF)} applies the same similarity filtering method to each instance, retaining only the top $(100-k)\%$ generations for each query.
% \end{itemize}
% We report results on both raw and filtered outputs as robustness checks.
% Full details are provided in Appendix~\ref{app:filter}.

\paragraph{Post-processing.}
Patchscopes generations from shallow layers can be noisy due to insufficient contextual priors, which may spuriously overlap with entities of interest.
We therefore apply global and local filtering based on sentence similarity, and report both raw and filtered outputs as robustness checks.
Full details are provided in Appendix~\ref{app:filter}.

\subsection{Causal Intervention}
\label{sec:causal intervertion}

To examine the causal role of knowledge recall at different token positions, we design a causal intervention method. 
The intervention consists of three runs:
\textbf{Clean Run}, where the model processes the original multi-hop query and produces the probability of the correct answer $P_{\mathrm{clean}}[o]$.
\textbf{Corrupted Run}, where Gaussian noise is added to the subject embedding, yielding $P_{\mathrm{corr}}[o]$.
\textbf{Patched Run}, where the activation at token position $i$ and layer $l$ in the clean run is replaced with its corrupted counterpart, resulting in $P_{\mathrm{patch}}^{(i,l)}[o]$.
By corrupting a specified activation in the clean run and propagating its effect forward, we estimate how much this activation contributes to the original prediction while preserving the original reasoning trajectory.
We measure the intervention effect as:
\[
\mathrm{IE}(i,l) = P_{\mathrm{clean}}[o] - P_{\mathrm{patch}}^{(i,l)}[o],
\]
where a larger value indicates a stronger causal contribution to the correct prediction.
Appendix~\ref{app:causal_intervention} further explains this design through a probabilistic-path derivation.

\section{Do LLMs Rely on Explicit Hop-aligned Circuits?}
\label{Do LLMs Rely on Explicit Multi-hop Reasoning Circuits?}

We decompose \textbf{RQ1} into two sub-questions:
\textit{(RQ1.1) Do LLMs exhibit latent reasoning when answering multi-hop queries?}
\textit{(RQ1.2) Is such reasoning implemented through hop-aligned circuits?}

\subsection{Probing Setup}
Following prior work~\cite{biran2024hopping-too-late, yao-etal-2025-cake}, we apply Patchscopes to the \textbf{Correct} subset by probing hidden states at the \textit{last subject token} and the \textit{last token}, and report (i) the decoding frequency of each entity and (ii) the earliest layer at which each entity becomes decodable.
% \hl{where both the multi-hop question and all associated single-hop questions are answered correctly.}

\subsection{Results}
% Figure~\ref{fig:patchscopes_class1} visualizes Patchscopes decoding results. Each subgraph reports the proportion of entities decoded from a given token across layers (bars), together with the average earliest layer at which each entity first occurs (lines). Bars of the same color correspond to multi-hop queries with the same number of hops, with entities ordered from left to right along the reasoning chain. Based on these results, we draw the following conclusions.

Figure~\ref{fig:patchscopes_class1} reports entity decoding proportions (bars) and the average earliest decodable layer (lines) for a given token. Bars of the same color indicate queries with the same total hops, and entities are ordered from left to right by hop index. Based on these results, we draw the following conclusions.

\textbf{RQ1.1: LLMs generally exhibit latent reasoning in multi-hop question answering.}
When probing the \textit{subject} token, all entities are frequently decodable, with the first bridge entity \(e_1\) often achieving the highest decoding rate (Figure~\ref{fig:patchscopes_class1} (a,d)).
Under filtered settings (Figure~\ref{fig:patchscopes_class1} (b-c,e-f)), the earliest decodable layer of entities tends to shift deeper as the hop index increases, consistent with progressive information accumulation across layers.
When probing the last token (Figure~\ref{fig:patchscopes_class1} (g-l)), bridge entities remain decodable, whereas the subject entity $e_0$ is rarely decoded.
These observations suggest that LLMs retain intermediate factual information in their hidden states during multi-hop reasoning.

\textbf{RQ1.2: However, Latent multi-hop reasoning does not consistently follow explicit hop-aligned circuits.}
At the last token, the final-hop entity is decoded with higher probability than earlier-hop entities, and its earliest decodable layer often shifts earlier as the hop index increases. This contrasts with hop-aligned circuit hypotheses, which predict similar decoding probabilities for bridge entities and monotonically increasing earliest decodable layers with hop index.

% Crucially, Table~\ref{tab:earliest_layer_raw} (see also Appendix~\ref{app: Average Earliest Layer}) shows that the last-hop entity at the last token (\textbf{bold}) often becomes decodable as early as, or earlier than, bridge entities at the subject token. We term this phenomenon \emph{layer-order inversion}. For example, on 4-hop queries in Llama~3, the final entity $e_4$ at the last token emerges earlier than the first bridge entity $e_1$ at the subject across all settings. Such behavior is incompatible with explicit hop-aligned circuits, which assume that bridge entities emerge sequentially across layers.

Crucially, Table~\ref{tab:earliest_layer_raw} (see also Appendix~\ref{app: Average Earliest Layer}) shows that the last-hop entity at the last token (\textbf{bold}) often becomes decodable as early as, or earlier than, bridge entities at the subject token. We term this phenomenon \emph{layer-order inversion}. For example, on 4-hop queries in Llama~3, the final entity $e_4$ at the last token emerges earlier than the first bridge entity $e_1$ at the subject across all settings. 
This ordering is inconsistent with explicit hop-aligned circuits, as this hypothesis requires bridge entities to become decodable before their dependent later-hop entities.
Further instance-level statistics in Table~\ref{tab:instance_inversion} in Appendix~\ref{app:instance_inversion} show that \emph{layer-order inversion} is not driven by a few extreme instances and becomes more prevalent as the number of hops increases.

\begin{figure*}[th]
    \centering
    \includegraphics[width=\textwidth]{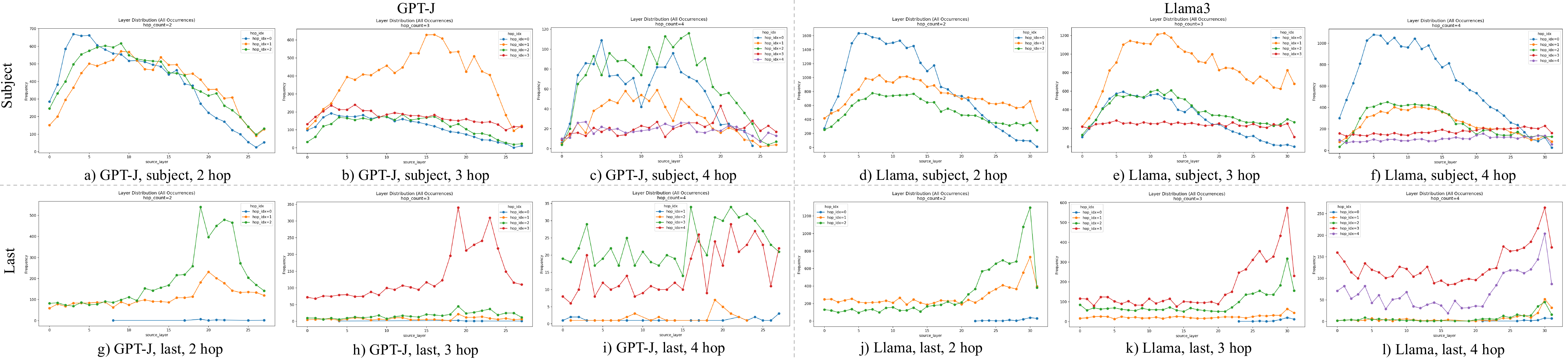}
    \caption{Layer-wise generation distributions produced by Patchscopes on the Correct subset. }
    \label{fig:Patchscopes distributions}
\end{figure*}

\begin{figure*}[th]
    \centering
    \includegraphics[width=\textwidth]{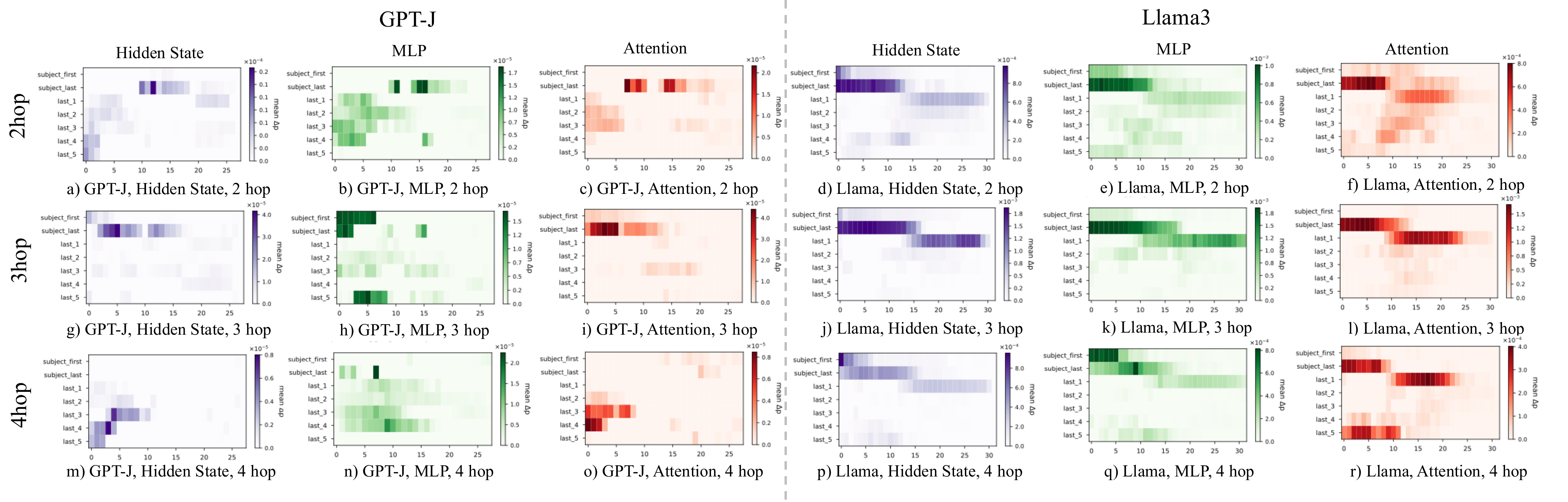}
    \caption{Causal intervention effects across layers and token positions on the Correct subset.}
    \label{fig:causal_inversion}
\end{figure*}

\subsection{Discussion}

A key reason prior work attributes multi-hop reasoning to hop-aligned circuits is the limited experimental scope. As discussed in Section~\ref{Multi-hop Reasoning Circuits}, most analyses focus on two-hop reasoning over synthetic, template-based datasets, which restricts observable latent behaviors. Consistently, we observe no layer-order inversion in two-hop queries, while the effect becomes more pronounced as hop count increases.

% At a deeper level, this discrepancy reflects a mismatch between the probabilistic nature of LLMs and symbolic interpretations of their internal computations. Prior analyses often interpret the most prominent internal activation patterns as evidence of deterministic, step-by-step reasoning.
% A similar misinterpretation also arose in the early development of quantum mechanics: as shown in Figure~\ref{fig:framework}(a), early atomic models assumed that electron orbits were fixed, but in reality they are probability clouds. By analogy, latent multi-hop reasoning in LLMs should be understood as a probabilistic recall process rather than a strictly symbolic computation, a perspective we develop in the next section.

At a deeper level, this discrepancy reflects a mismatch between the probabilistic nature of LLMs and symbolic interpretations of their internal computations. Prior analyses often interpret the most prominent internal activation patterns as evidence of deterministic, step-by-step reasoning.
A similar misinterpretation also appeared in early atomic models, which assumed fixed electron orbits rather than probability clouds (Figure~\ref{fig:framework}(a)). 
By analogy, rather than a strictly symbolic computation, latent multi-hop reasoning in LLMs should be understood as a probabilistic recall process, which we investigate systematically in the next section.

\section{A Probabilistic Perspective on the Recall-and-Extract Framework}
\label{Sec: A Probabilistic Perspective on the Recall-and-Extract Framework}

To answer \textbf{RQ2}, we extend the Recall-and-Extract framework~\cite{geva2023dissecting-recall-then-extract} from a probabilistic perspective by introducing horizontal recall and intuition-like recall.
We focus on two sub-questions:
\textit{(RQ2.1) How LLMs probabilistically acquire the knowledge for multi-hop reasoning?} and
\textit{(RQ2.2) How can such a probabilistic mechanism account for prior empirical observations?}

\subsection{Probing Setup}

To address \textbf{RQ2.1}, we conduct three complementary analyses on the \textbf{Correct} subset. Specifically,
(1) we analyze Patchscopes across layers to characterize the probabilistic distribution of entity-level knowledge in intermediate hidden states;
(2) causal intervention analysis is conducted to assess causal effects at the subject token and the last $k$ tokens ($k=5$) across hidden states, MLP outputs, and attention outputs;
(3) we further perform hidden-state similarity analysis as supplementary evidence to track how internal representations evolve across layers, with details provided in Appendix~\ref{sec:hidden_state_similarity}.

\subsection{Results}
\label{Mechanism}

Figure~\ref{fig:Patchscopes distributions} illustrates the distribution of entities decoded by Patchscopes across layers, where curves of different colors correspond to entities at different hop indexes.
Figure~\ref{fig:causal_inversion} reports causal intervention effects across layers and token positions, with darker colors indicating larger decreases in the correct-answer probability after intervention.
Figure~\ref{fig:sim_mlp_fc_in_sub_sub} in Appendix~\ref{sec:hidden_state_similarity} provides hidden-state similarity results between multi-hop queries and their corresponding single-hop queries, grouped by \textit{(hop index -- total hops)}.
Together, these results suggest a probabilistic extension of the Recall-and-Extract mechanism, which we summarize below.

% \textbf{RQ2.1: Latent multi-hop reasoning in LLMs is supported by a probabilistic recall-and-extract mechanism}, in which MLP layers perform probabilistic knowledge recall conditioned on both the current hidden state (vertical) and previously processed tokens (horizontal), while higher-layer attention amplifies the probability of answer-relevant knowledge based on the query, as illustrated in Figure~\ref{fig:framework}(d).

\textbf{RQ2.1: Latent multi-hop reasoning in LLMs is supported by a probabilistic recall-and-extract mechanism,} which can be summarized by three key points as shown in  Figure~\ref{fig:framework}(d).

\paragraph{Probabilistic recall operates both vertically and horizontally.}
Our results suggest that knowledge recall in LLMs is jointly shaped by vertical recall across layers and horizontal recall across tokens.
Vertical recall has been suggested in prior studies~\cite{meng2022locating-rome,geva2023dissecting-recall-then-extract}, where MLP layers progressively recall factual knowledge conditioned on the current representation.
Consistent with this view, Figure~\ref{fig:Patchscopes distributions} (a--f) shows that entities from different hop indexes can be decoded across layers at the subject position, while Figure~\ref{fig:sim_mlp_fc_in_sub_sub} provides supplementary evidence that representations gradually shift from earlier-hop to later-hop knowledge.
However, vertical recall alone cannot explain \emph{layer-order inversion}. 
We therefore propose horizontal recall, where accumulated context guides knowledge recall.
As shown in Figure~\ref{fig:Patchscopes distributions}(g--j), later-hop entities are often more decodable than earlier-hop entities at the last token even in shallow layers, indicating that deeper-hop knowledge can become available without explicitly decoding all preceding bridge entities, a behavior consistent with the autoregressive next-token prediction objective of LLMs~\cite{brown2020language-gpt3}.

Notably, vertical and horizontal recall can interact: early-hop knowledge obtained through vertical recall can increase the probability of horizontal deeper-hop recall, and vice versa. Such coupling jointly shapes how knowledge distributions evolve during latent multi-hop reasoning~\cite{zhangscaling-law-llm, wu2025inference-scaling-law}.

\paragraph{Last-token shallow recall acts as an intuition-like signal.}
Horizontal recall explains how deeper-hop knowledge becomes available at the last token, but its causal effect on prediction remains unclear.
To examine its causal role, we conduct the causal intervention analysis.
As shown in Figure~\ref{fig:causal_inversion}, for the weaker GPT-J model, intervening on shallow activations at the last tokens produces stronger effects than interventions at the subject token, with the effect increasing in higher-hop queries.
In contrast, for stronger Llama3-8B model, last-token interventions are much weaker than subject.

These results suggest that shallow horizontal recall of deeper-hop knowledge at the last token behaves as an intuition-like signal for answer selection.
Analogous to human intuition, it may not alter the reasoning process itself, but can bias the final judgment.
In LLMs, attention masking prevents this signal from influencing earlier tokens, so it does not directly participate in preceding hop-aligned recall.
Nevertheless, it can still affect the final prediction, with its influence increasing for weaker models and higher-hop queries.

% \paragraph{Attention as knowledge extraction.}
% An interesting phenomenon can also be observed in Figure~\ref{fig:Patchscopes distributions}(g-j), the decoding probability of the answer entity at the last token increases sharply in deeper layers. This indicates that answer-relevant information is selectively amplified toward the end of the network.
% This empirically validates the Recall-and-Extract view that deeper-layer attention plays a key role in knowledge extraction~\cite{geva2023dissecting-recall-then-extract, li2024pmet, tamayo2024mass-editing-memory-with-attention}.

\paragraph{Attention extracts answer-relevant knowledge.}
Figure~\ref{fig:Patchscopes distributions}(g--j) shows a sharp increase in answer-entity decodability at the last token in deeper layers, aligning with the Recall-and-Extract view that deeper-layer attention extracts answer-relevant information~\cite{geva2023dissecting-recall-then-extract, li2024pmet, tamayo2024mass-editing-memory-with-attention}.
Causal intervention further supports this view, as perturbing deep-layer attention at the last token has clear effects that grow with hop count (Figure~\ref{fig:causal_inversion}(f,l,r)).

\subsection{Discussion}
\noindent\textbf{RQ2.2: The probabilistic recall-and-extract framework provides a more complete and coherent account of prior multi-hop reasoning phenomena.} Its advantages are most evident in explaining hop-aligned circuits and shortcut.

\paragraph{Hop-aligned Circuits Hypothesis.}

Prior work~\cite{yao-etal-2025-cake} interprets layer-wise decoding patterns as evidence that LLMs resolve each hop sequentially. Under our probabilistic recall-and-extract framework, these observations are more naturally explained as reflecting the most probable recall trajectory across token positions. For example, at the subject position, the probability of decoding the first bridge entity \(e_1\) is typically higher than that of decoding later-hop entities (e.g., \(e_2\)) at the last token, making the ``\(e_1 \rightarrow e_2\)'' trajectory more likely to be observed in layer-wise analyses. Crucially, this probabilistic interpretation not only accounts for such circuit-like patterns in restricted settings (e.g., two-hop), but also accommodates higher-hop behaviors such as \emph{layer-order inversion}, which are difficult to reconcile with strictly hop-aligned circuit explanations.

\begin{figure*}[th]
    \centering
    \includegraphics[width=\textwidth]{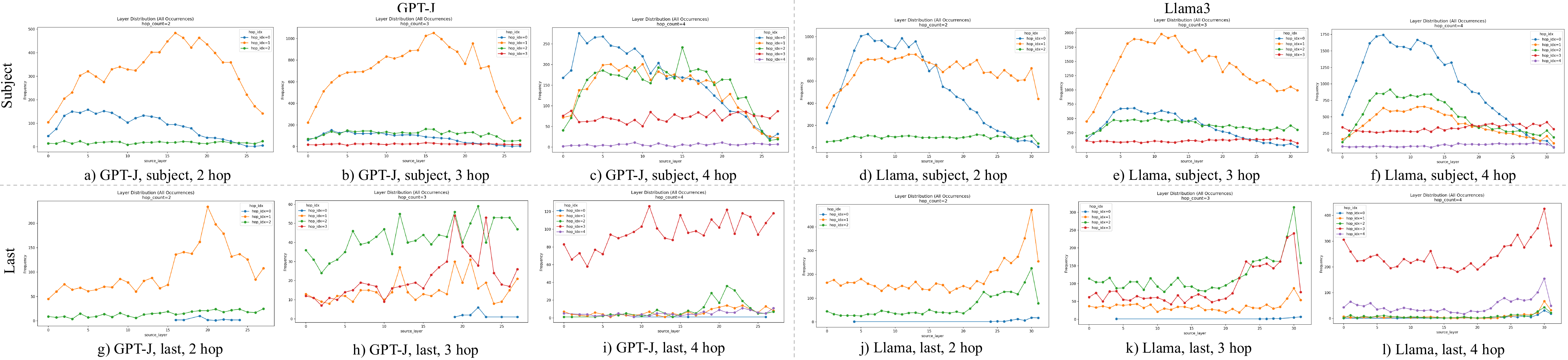}
    \caption{Layer-wise generation distributions produced by Patchscopes on the Incorrect subset. }
    \label{fig:Patchscopes distributions class3}
\end{figure*}

\paragraph{Shortcut.}
Shortcuts~\cite{ju2024investigating-Multi-Hop-Factual-Shortcuts, zhang2024enhancing-multi-hop-reasoning-knowledge-erasure} occur when an LLM answers a multi-hop question correctly despite failing on some corresponding single-hop questions.
Under our framework, shortcuts can be viewed as cases where the intuition-like signal is directly used for answer selection.
Our causal intervention results suggest that such direct recall paths become more influential for weaker models or harder queries, where reliable hop-by-hop reasoning is less likely to dominate.
This aligns with broader shortcut-learning studies showing that models may exploit shallow associative features instead of robust reasoning in complex scenarios~\cite{geirhos2020shortcut-in-deep-neural-networks-nmi,yuan-etal-2024-llms-shortcut}.

Beyond the above phenomena, knockout~\cite{geva2023dissecting-recall-then-extract}, back-patching~\cite{biran2024hopping-too-late}, and Chain-of-Thought~\cite{wei2022chain-of-thought} are also consistent with our framework, as they modulate shallow-layer recall of deeper-hop knowledge through suppression, restoration, or contextual enrichment.
Detailed analyses are provided in Appendix~\ref{app:analyses}.

% \begin{figure*}[th]
%     \centering
%     \includegraphics[width=\textwidth]{patchscopes_layer_distribution_class3.pdf}
%     \caption{Layer-wise generation distributions produced by Patchscopes on the Incorrect subset. }
%     \label{fig:Patchscopes distributions class3}
% \end{figure*}

\section{Reinterpreting Errors in Multi-hop Reasoning}
\label{Sec: Reinterpreting Errors in Multi-hop Reasoning}
In this section, we address \textbf{RQ3} by reinterpreting errors in multi-hop reasoning under our probabilistic recall-and-extract framework, providing new insights for model intervention and editing.

\subsection{Probing Setup}
To investigate \textbf{RQ3}, we apply Patchscopes to instances in the \textbf{Incorrect} and \textbf{Missing} subsets. 
In addition, we further compare layer-wise hidden-state similarity between \textbf{Correct} and \textbf{Incorrect} to isolate internal differences when single-hop facts are correct but multi-hop reasoning fails.

\subsection{Results}
We first focus on \textbf{Incorrect} cases, where all constituent single-hop facts are answered correctly, allowing us to analyze multi-hop reasoning failures without confounding effects from missing prerequisite knowledge. We attribute the observed failures to two main factors.

\textbf{RQ3.1: Insufficient shallow recall of deeper-hop knowledge.}
Figure~\ref{fig:Patchscopes distributions class3} shows that deeper-hop knowledge is insufficiently decoded in shallow layers, especially at the last token.
This indicates that the model fails to form sufficient shallow recall of deeper-hop entities under the multi-hop context.
Figure~\ref{fig:patchscopes_class3} in Appendix~\ref{app:incorrect_patchscopes} further shows lower overall decoding rates of deeper-hop entities in \textbf{Incorrect} cases.
Moreover, hidden-state similarity analysis in Figure~\ref{fig:sim_mlp_fc_in_sub_sub} in Appendix~\ref{sec:hidden_state_similarity} shows that the gap between \textbf{Correct} and \textbf{Incorrect} cases increases as hop depth grows.
These results suggest that even when the model answers all corresponding single-hop questions correctly, it may fail to activate sufficient deeper-hop knowledge under different multi-hop contexts, a limitation also noted in prior work~\cite{yang-etal-2025-latent-multi-hop-reasoning-wo-shortcuts}.

\textbf{RQ3.2: Insufficient knowledge extraction in deeper attention layers.}
As shown in Figure~\ref{fig:Patchscopes distributions class3}(g-j), even at the last token in deeper layers, the final-hop entity often fails to dominate the decoding distribution: its probability may remain low or only comparable to that of bridge entities.
This suggests that deeper-layer attention may not sufficiently amplify answer-relevant knowledge to support accurate generation, resulting in extraction failures despite partial recall.

We further analyze \textbf{Missing} cases in Appendix~\ref{app: missing}. When single-hop facts are answered incorrectly, early-hop knowledge (e.g., the first-hop entity) also exhibits low recall probability under the multi-hop query, preventing correct answer generation.

\subsection{Discussion}

Under our probabilistic recall-and-extract framework, both hop-aligned reasoning and intuition-like recall rely on context-conditioned knowledge recall.
Our results show that such recall differs between multi-hop and single-hop settings, even when they involve the same underlying knowledge.
This offers new insights for related research areas.
For example, in knowledge editing, key representations should move beyond subject-centric samples and be derived through additional techniques to obtain more universal representations~\cite{zhang2024knowledge-graph-ke}.
Moreover, attention-encoded knowledge deserves closer study, as existing methods~\cite{meng2023mass-memit,fang2025alphaedit} mainly focus on MLP-stored knowledge.

\section{Conclusion}
\label{sec:conclusion}

% In this work, we investigated latent multi-hop reasoning in large language models and identified \emph{layer-order inversion}, where later-hop entities become decodable earlier than early-hop bridge entities as hop count increases.
% This finding exposes fundamental limitations of hop-aligned circuit hypothesis.
% To reinterpret this observation, we proposed a \emph{probabilistic recall-and-extract} framework, which views latent reasoning as probabilistic knowledge recall guided jointly by current representations and accumulated context, followed by selective extraction in deeper layers. Under this perspective, hop-aligned circuits correspond to dominant high-probability recall paths, while chain-of-thought prompting acts by enriching contextual priors. Finally, we showed that failures in multi-hop reasoning arise from mismatches between multi-hop and single-hop knowledge recall and from insufficient extraction of knowledge. Future work will further clarify the knowledge encoded within the attention module and explore their implications for model intervention and knowledge editing.

In this work, we investigated latent multi-hop reasoning in LLMs and identified \emph{layer-order inversion}, where later-hop entities become decodable earlier than early-hop bridge entities as hop count increases.
This finding challenges the hop-aligned circuit hypothesis and motivates our \emph{probabilistic recall-and-extract} framework, which views latent reasoning as a superposition of probabilistic reasoning paths.
Under this framework, we introduce horizontal recall across tokens and intuition-like shallow recall at the last token.
Causal intervention further shows that such last-token recall can influence answer selection, with stronger effects in weaker models and higher-hop queries.
Our framework further interprets hop-aligned circuits as dominant high-probability recall paths, explains the role of horizontal recall in shortcut behaviors, and diagnoses multi-hop failures as insufficient recall or extraction.
In future work, we will further investigate how intuition-like recall can be strengthened to improve LLM reasoning performance.

\section*{Limitations}
Firstly, constrained by computational resources, our experiments are conducted on relatively small-scale models. While we are unable to evaluate larger models (e.g., 70B parameters), we believe that the observed phenomena are sufficient to reveal limitations of existing hop-aligned circuit hypotheses. 
Rather than claiming universality across model scales, our goal is to motivate a probabilistic perspective on multi-hop reasoning that better accounts for the observed behaviors.

Second, our interpretation of intuition-like recall signals remains hypothesis-driven.
It is motivated by Patchscopes results and causal intervention analysis, which suggest that shallow horizontal recall of deeper-hop knowledge at the last token can influence answer selection.
However, we do not further verify how such signals can be guided, strengthened, or incorporated into inference strategies to improve LLM reasoning performance.
Since intuition-like recall may interact with step-by-step reasoning under the probabilistic recall framework, designing effective methods to leverage this signal requires further theoretical analysis and controlled experiments.
We leave this direction for future work.
Nevertheless, intuition-like recall offers a new perspective for understanding and improving latent multi-hop reasoning in LLMs.

\section*{Acknowledgements}
This research was supported by grants from the National Key Research and Development Program of China (Grant No. 2024YFC3308200), the National Natural Science Foundation of China (Grant Nos. U25B2072 and 62406303), the Key Technologies R \& D Program of Anhui Province (No. 202423k09020039).

\bibliography{custom}

\appendix

\section{Supplementary Experiment}

\subsection{Patchscopes Analysis}
\label{app:filter}

Due to space constraints, we provide an overview of Patchscopes in Section~\ref{sec:patchscopes}. Here, we present the detailed implementation.

\subsubsection{Setting}
We adopt Patchscopes~\cite{ghandeharioun2024patchscopes} as the primary tool to probe latent multi-hop reasoning in LLMs. 
Patchscopes analyzes internal knowledge by extracting the hidden state from a selected token at a given layer when processing the original query, injecting it into a placeholder token \textit{x} in an explanatory target prompt like \textit{Syria: Syria is a country in the Middle East, Leonardo DiCaprio: Leonardo DiCaprio is an American actor, Samsung: Samsung is a South Korean multinational corporation, x}, and observing the generation. 
The generated text provides a natural language interpretation of the knowledge encoded in the source hidden state.
Compared to probing methods~\cite{belinkov2022probing, belrose2023eliciting-tuned-lens}, Patchscopes is fully unsupervised and enables direct visualization of knowledge encoded in earlier layers.

Following prior work~\cite{biran2024hopping-too-late}, we repeat the patching procedure three times per query and report aggregated statistics. 
Despite using a different dataset, we observe consistent trends on two-hop queries with previous work, supporting the reliability of our results.

\subsubsection{Post-processing and Filtering.}
When decoding from shallow layers, Patchscopes may produce random outputs due to insufficient contextual priors. 
Such generations can still include frequent entities (e.g., countries or cities), which may coincidentally overlap with bridge entities and introduce noise when estimating the earliest emergence layer.
To mitigate this issue, we try three filters:

\begin{itemize}[leftmargin=*, itemsep=0pt]
    \item \textbf{Global filtering (GF)} computes the similarity~\cite{reimers2019sentence-bert} between generated outputs and the original query, and removes the bottom $k\%$ of results with the lowest similarity across the entire dataset.
    \item \textbf{Local filtering (LF)} applies the same similarity filtering method to each instance, retaining only the top $(100-k)\%$ generations for each query.
    \item \textbf{Layer filtering} discards generations originating from source layers below a predefined threshold.
\end{itemize}

\begin{figure*}[!ht]
    \centering
    \includegraphics[width=\textwidth]{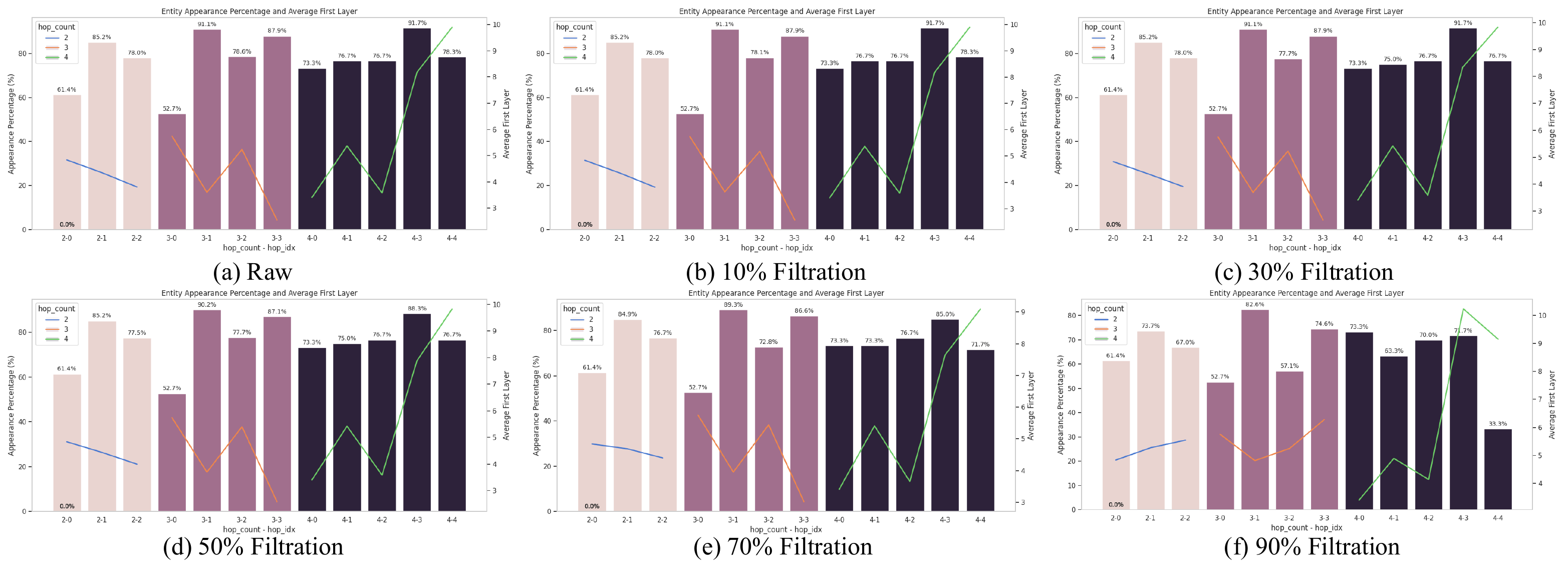}
    \caption{Patchscopes decoding results at subject token on GPT-J-6B under different global filtering ratios.}
    \label{fig:app_global_filter}
\end{figure*}

\begin{figure*}[!ht]
    \centering
    \includegraphics[width=\textwidth]{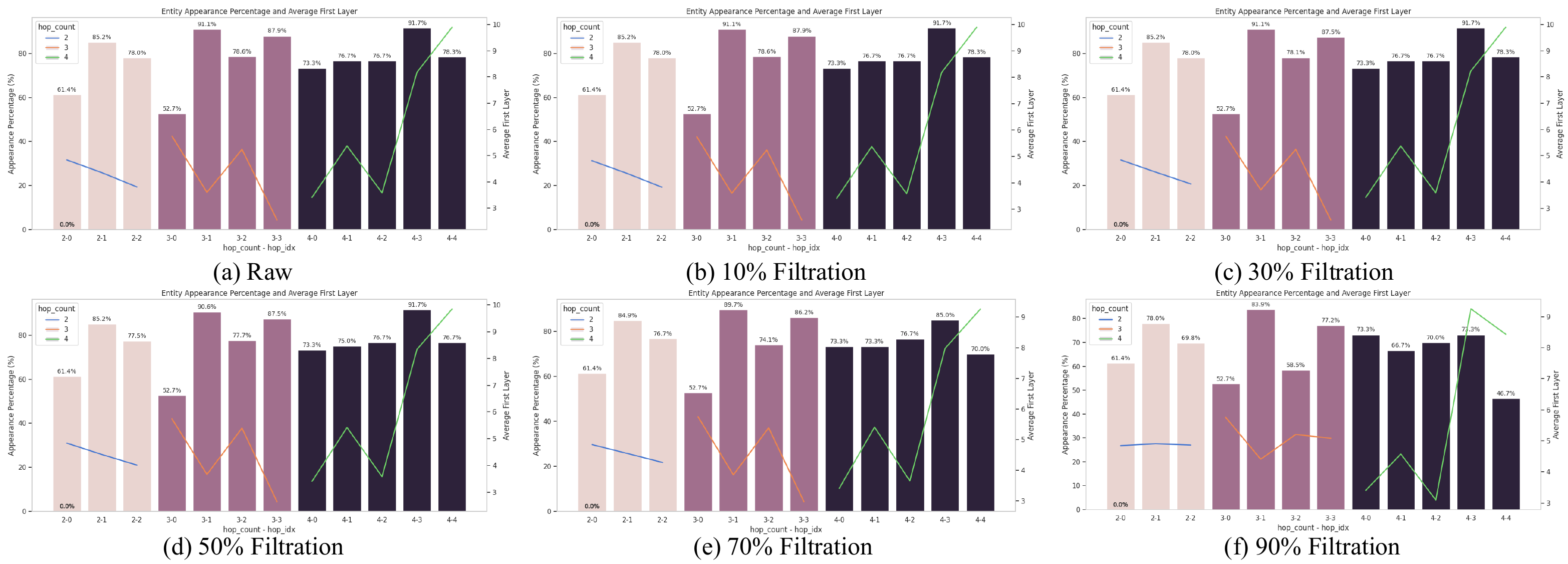}
    \caption{Patchscopes decoding results at subject token on GPT-J-6B under different local filtering ratios.}
    \label{fig:app_case_filter}
\end{figure*}

\begin{figure*}[!ht]
    \centering
    \includegraphics[width=\textwidth]{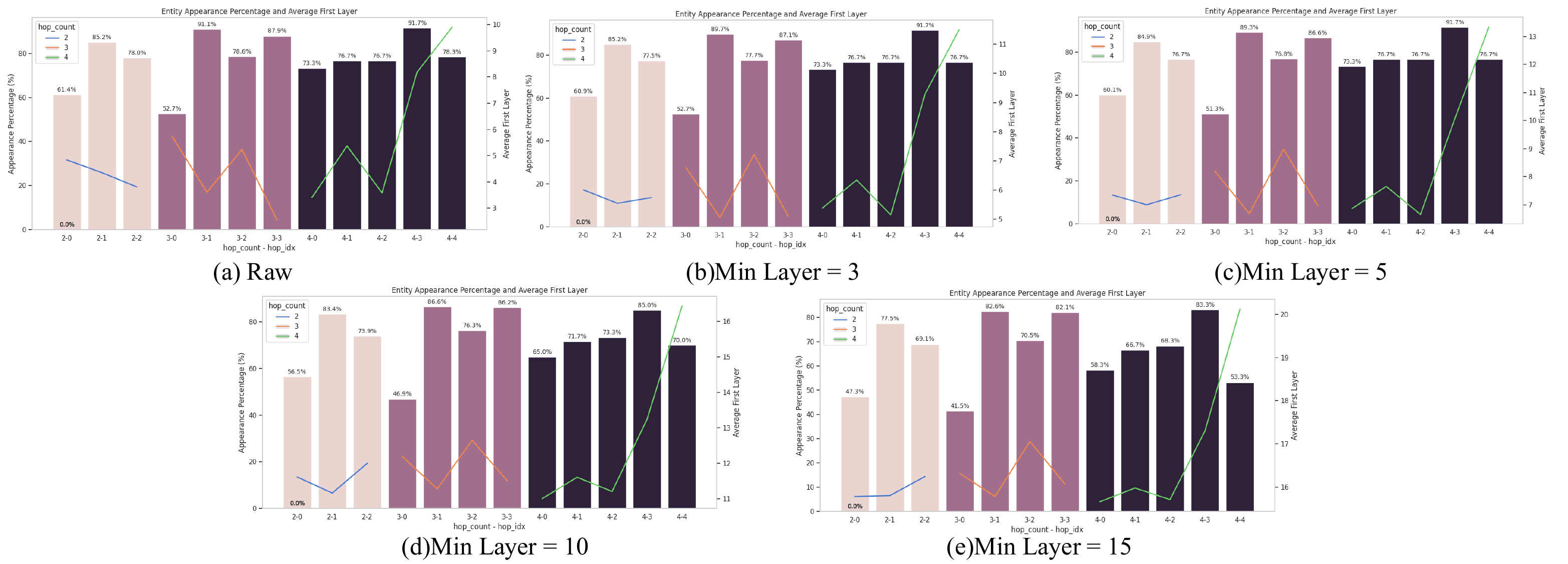}
    \caption{Patchscopes decoding results at subject token on GPT-J-6B with different minimum layer.}
    \label{fig:app_minlayer}
\end{figure*}

As shown in the Figures~\ref{fig:app_case_filter},~\ref{fig:app_global_filter}, and~\ref{fig:app_minlayer}, the overall decoding distributions remain largely unchanged under most filtering, and noticeable shifts only emerge under strong filtering, specifically with 90\% global filtering and 90\% local filtering. We therefore focus on these two settings in the main analysis. 
We therefore report results on the raw Patchscopes outputs as well as those obtained with 90\% global and local filtering.

\begin{figure}[!t]
    \centering
    \includegraphics[width=\linewidth]{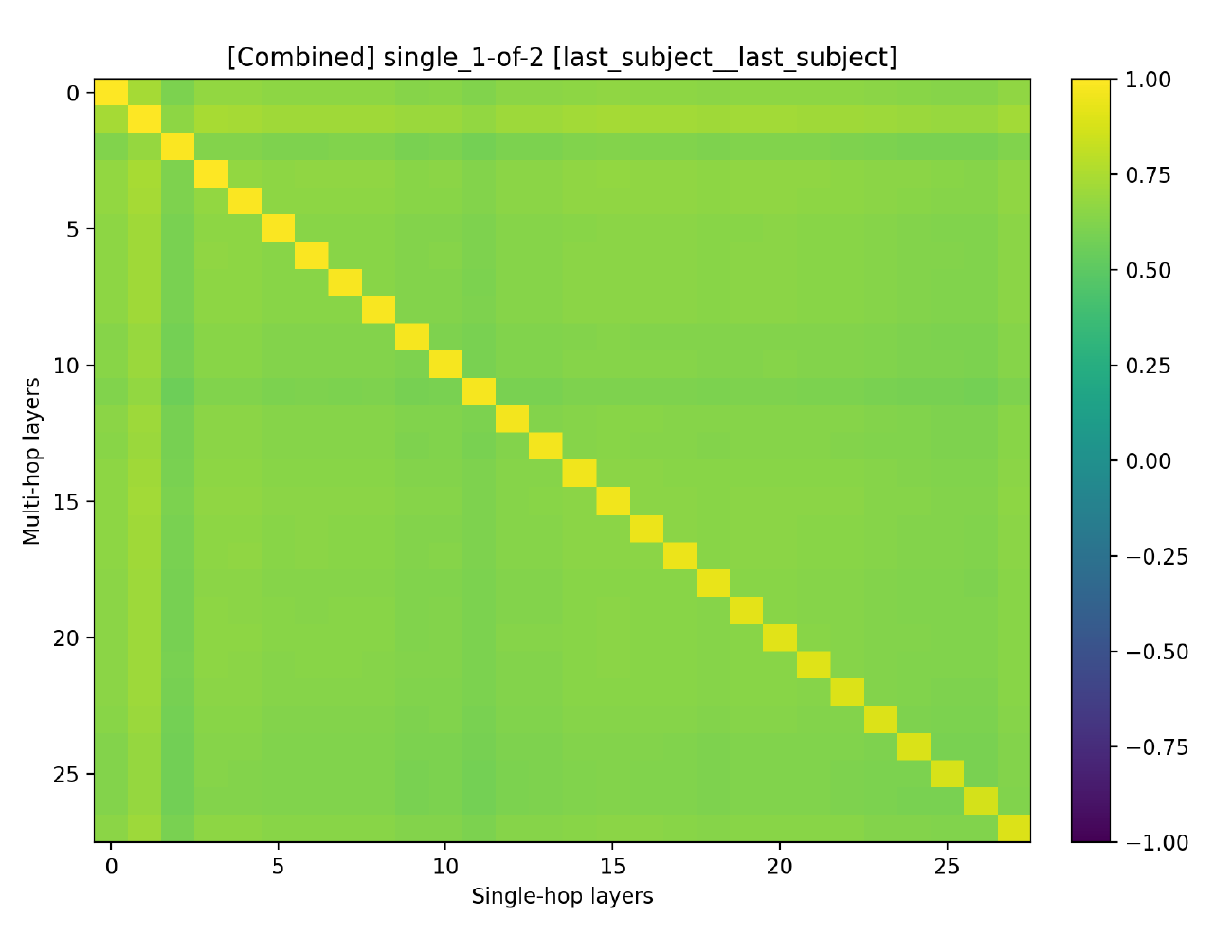}
    \caption{Cross-layer Similarity of Subject \textit{mlp\_fc\_in} Representations on GPT-J-6B.}
    \label{fig:cross_layer_sim}
\end{figure}

\begin{figure*}[th]
    \centering
    \includegraphics[width=\textwidth]{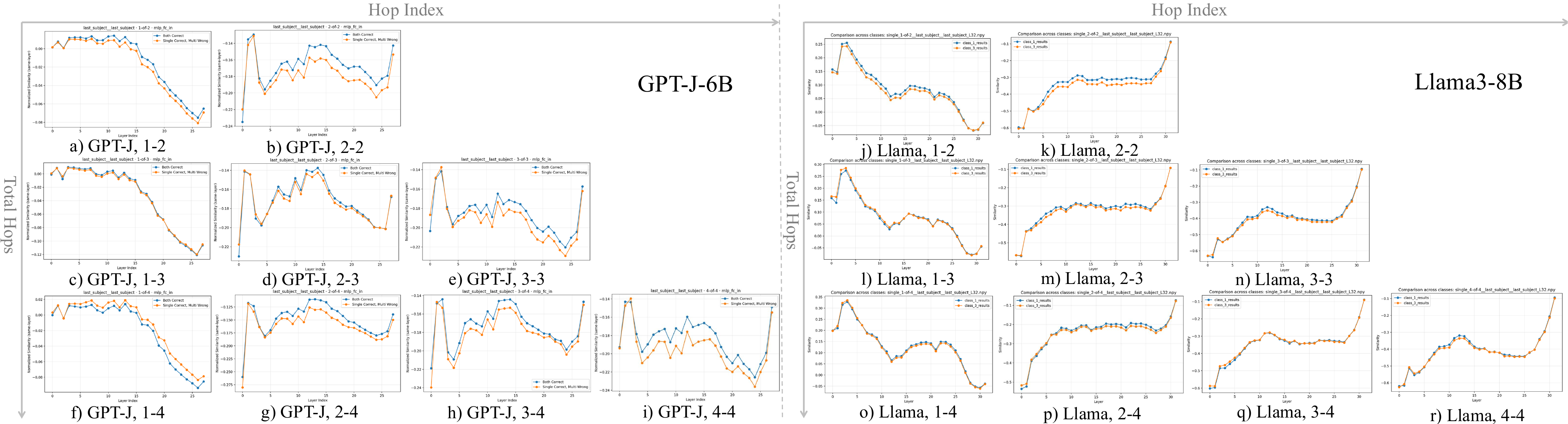}
    \caption{Normalized similarity of \textit{mlp\_fc\_in} at the subject between multi-hop queries and their single-hop queries. }
    \label{fig:sim_mlp_fc_in_sub_sub}
\end{figure*}

\begin{figure*}[th]
    \centering
    \includegraphics[width=\textwidth]{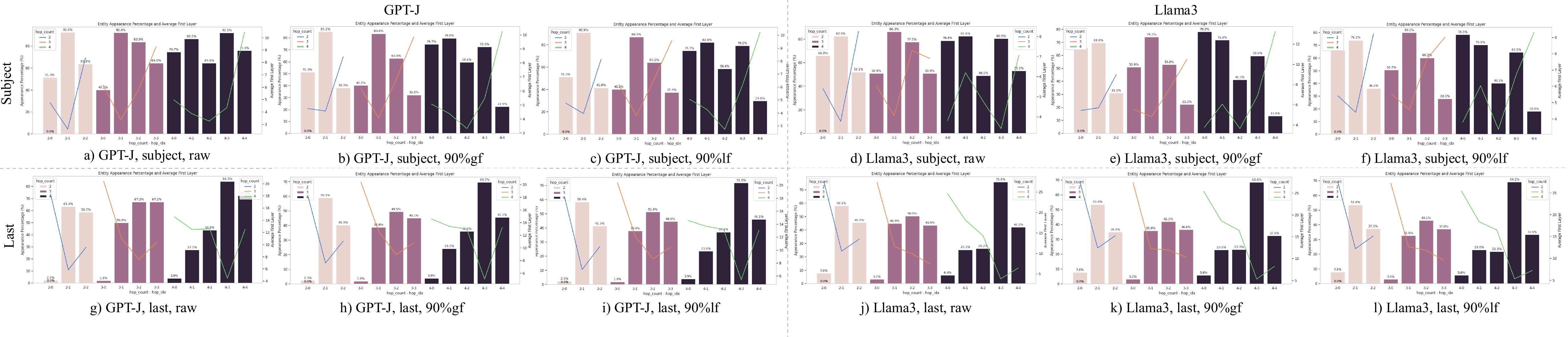}
    \caption{Patchscopes results for GPT-J (left) and Llama3 (right) on the Incorrect subset.}
    \label{fig:patchscopes_class3}
\end{figure*}

\subsection{Causal Intervention}
\label{app:causal_intervention}

In Section~\ref{sec:causal intervertion}, we introduce a causal intervention method to analyze the role of internal representations. 
Different from prior restoration-based approaches~\cite{meng2022locating-rome,yao-etal-2025-language-models-multi-hop-reasoning-training-data}, we apply corruption to the clean run rather than restoring clean activations in a corrupted run.

The key motivation comes from our probabilistic recall-and-extract framework in Section~\ref{Sec: A Probabilistic Perspective on the Recall-and-Extract Framework}, where multi-hop reasoning is modeled as a superposition of reasoning paths:

\begin{equation}
    P(e_n \mid x) = \sum_{\pi \in \Pi} P(\pi \mid x)\,P(e_n \mid \pi, x).
\end{equation}

We decompose the prediction into two dominant paths, 
the hop-aligned reasoning path $\pi_{\text{hop}}$ and the direct recall path $\pi_{\text{direct}}$:

\begin{equation}
\begin{aligned}
P(e_n \mid x) \approx\;&
P(\pi_{\text{hop}} \mid x)\!\!\prod_{i=1}^{n} P(e_i \mid e_{i-1}, r_i) \\
&+ P(\pi_{\text{direct}} \mid x)\,P(e_n \mid \pi_{\text{direct}}, x) \\
&+ \sum_{\pi \in \Pi'} P(\pi \mid x)\,P(e_n \mid \pi, x).
\end{aligned}
\end{equation}

This decomposition shows that the final prediction depends not only on recalled information, but also on the underlying path distribution $P(\pi \mid x)$, which is determined by preceding tokens. 
Therefore, perturbing the input (e.g., subject embedding) can disrupt the entire reasoning path, making it difficult to isolate the contribution of later representations.

To preserve the reasoning paths while isolating local effects, we apply corrupted activations to the clean run. 
This allows us to better measure how specific representations, especially at the last token, affect the final prediction without altering the overall reasoning trajectory.

\begin{table*}[t]
\centering
\small
\begin{tabular}{c c|ccc|cccc|ccccc}
\toprule
\multirow{2}{*}{Model} & \multirow{2}{*}{Token}
& \multicolumn{3}{c}{2-hop}
& \multicolumn{4}{c}{3-hop}
& \multicolumn{5}{c}{4-hop} \\
\cmidrule(lr){3-5} \cmidrule(lr){6-9} \cmidrule(lr){10-14}
& & $e_0$ & $e_1$ & $e_2$
  & $e_0$ & $e_1$ & $e_2$ & $e_3$
  & $e_0$ & $e_1$ & $e_2$ & $e_3$ & $e_4$ \\
\midrule
\multirow{2}{*}{GPT-J}
& Subject
& 4.84 & 5.27 & 5.55
& 5.75 & 5.25 & 5.25 & 6.28
& 3.41 & 4.89 & 4.14 & 10.23 & 9.15 \\
& Last
& 19.36 & 10.12 & \textbf{9.62}
& 18.57 & 11.68 & 13.43 & \textbf{6.38}
& 20.00 & 13.00 & 12.21 & 11.74 & \textbf{11.93} \\
\midrule
\multirow{2}{*}{Llama3}
& Subject
& 4.73 & 6.34 & 6.98
& 5.34 & 4.92 & 7.01 & 8.86
& 4.40 & 5.93 & 3.33 & 6.94 & 10.26 \\
& Last
& 28.92 & 13.21 & \textbf{12.20}
& 29.23 & 14.68 & 17.32 & \textbf{7.84}
& 27.88 & 21.00 & 15.39 & 6.96 & \textbf{5.76} \\
\bottomrule
\end{tabular}
\caption{Average Earliest Layer of Entity Emergence under 90\% Global Filter}
\label{tab:earliest_layer_gf}
\end{table*}

\begin{table*}[t]
\centering
\small
\begin{tabular}{c c|ccc|cccc|ccccc}
\toprule
\multirow{2}{*}{Model} & \multirow{2}{*}{Token}
& \multicolumn{3}{c}{2-hop}
& \multicolumn{4}{c}{3-hop}
& \multicolumn{5}{c}{4-hop} \\
\cmidrule(lr){3-5} \cmidrule(lr){6-9} \cmidrule(lr){10-14}
& & $e_0$ & $e_1$ & $e_2$
  & $e_0$ & $e_1$ & $e_2$ & $e_3$
  & $e_0$ & $e_1$ & $e_2$ & $e_3$ & $e_4$ \\
\midrule
\multirow{2}{*}{GPT-J}
& Subject
& 4.85 & 4.91 & 4.86
& 5.75 & 4.41 & 5.21 & 5.08
& 3.41 & 4.58 & 3.10 & 9.25 & 8.43 \\
& Last
& 19.36 & 10.37 & \textbf{9.52}
& 18.57 & 11.68 & 13.19 & \textbf{6.54}
& 20.00 & 13.00 & 12.50 & 11.19 & \textbf{11.57} \\
\midrule
\multirow{2}{*}{Llama3}
& Subject
& 4.73 & 5.43 & 5.83
& 5.42 & 4.67 & 6.68 & 6.79
& 4.28 & 5.88 & 2.54 & 6.36 & 6.98 \\
& Last
& 28.92 & 13.33 & \textbf{11.53}
& 29.23 & 15.87 & 16.94 & \textbf{7.44}
& 27.88 & 20.85 & 16.13 & 8.01 & \textbf{5.10} \\
\bottomrule
\end{tabular}
\caption{Average Earliest Layer of Entity Emergence under 90\% Local Filter}
\label{tab:earliest_layer_lf}
\end{table*}

\subsection{Hidden State Similarity}
\label{sec:hidden_state_similarity}

Section~\ref{Sec: A Probabilistic Perspective on the Recall-and-Extract Framework} reports hidden-state similarity experiments as supporting evidence. Due to space constraints, we describe the detailed implementation and additional results in this section.

\subsubsection{Setting}

We further conduct a hidden state similarity analysis~\cite{phang2021fine-similar-representations, jiang2024tracing-Layer-Wise-Similarity} as a supplementary experiment to examine how knowledge evolves during multi-hop reasoning. 
This analysis is motivated by prior work viewing MLPs in transformers as key–value memories for factual knowledge~\cite{geva2021transformer-key-value}.
Formally, the hidden state of token $i$ at layer $l$ is updated as
\begin{equation}
\begin{aligned}
h_i^{(l)} &= h_i^{(l-1)} + a_i^{(l)} + m_i^{(l)}, \\
a_i^{(l)} &= \mathrm{attn}^{(l)}\!\left(h_1^{(l-1)}, h_2^{(l-1)}, \dots, h_i^{(l-1)}\right), \\
m_i^{(l)} &= W_{\mathrm{proj}}^{(l)} \, \sigma\!\left(W_{\mathrm{fc}}^{(l)} \, \mathrm{LN}\!\left(a_i^{(l)} + h_i^{(l-1)}\right)\right).
\end{aligned}
\end{equation}
Based on this formulation, we track three representative hidden states: the attention projection output \textit{attn\_proj\_out}, corresponding to $a_i^{(l)}$, the MLP input \textit{mlp\_fc\_in}, serving as a key representation, and the MLP output \textit{mlp\_fc\_out}, corresponding to retrieved knowledge $m_i^{(l)}$.
We compute similarity scores between hidden states from a multi-hop query and those from its corresponding single-hop queries. 
As shown in Figure~\ref{fig:cross_layer_sim}, cross-layer similarities are substantially weaker than same-layer similarities, likely due to mismatched representation spaces. 
We therefore restrict our analysis to same-layer comparisons between key tokens.

\paragraph{Additional Results}

\begin{figure*}[!ht]
    \centering
    \includegraphics[width=\textwidth]{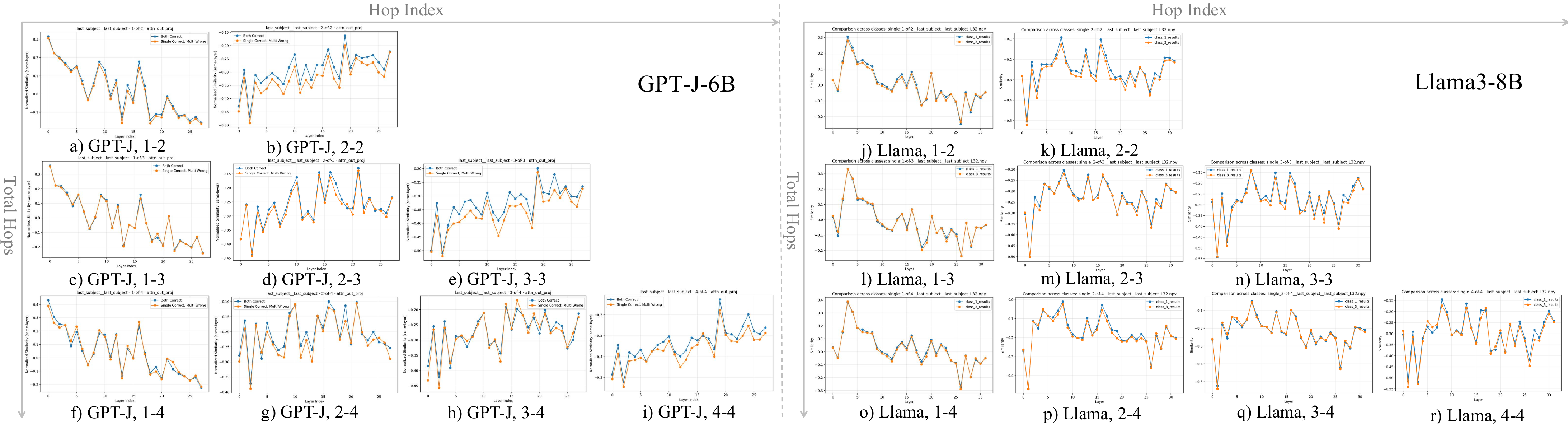}
    \caption{Normalized similarity of \textit{attn\_proj\_out} at the subject between multi-hop queries and their single-hop queries. }
    \label{fig:sim_attn_sub_sub}
\end{figure*}

\begin{figure*}[!ht]
    \centering
    \includegraphics[width=\textwidth]{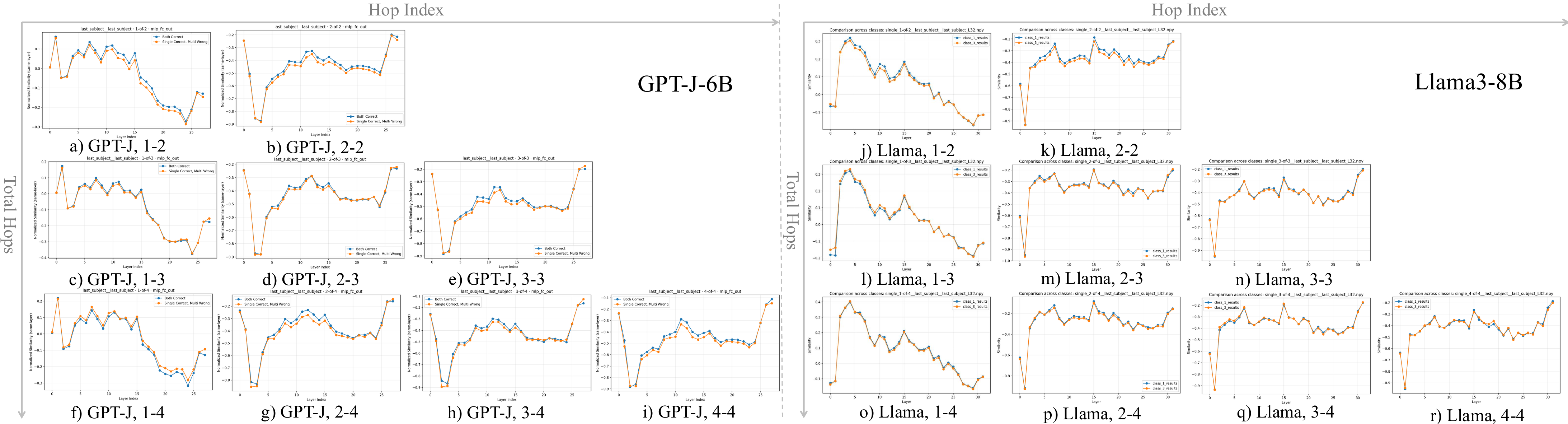}
    \caption{Normalized similarity of \textit{mlp\_fc\_out} at the subject between multi-hop queries and their single-hop queries. }
    \label{fig:sim_mlp_fc_out_sub_sub}
\end{figure*}

\begin{figure*}[!ht]
    \centering
    \includegraphics[width=\textwidth]{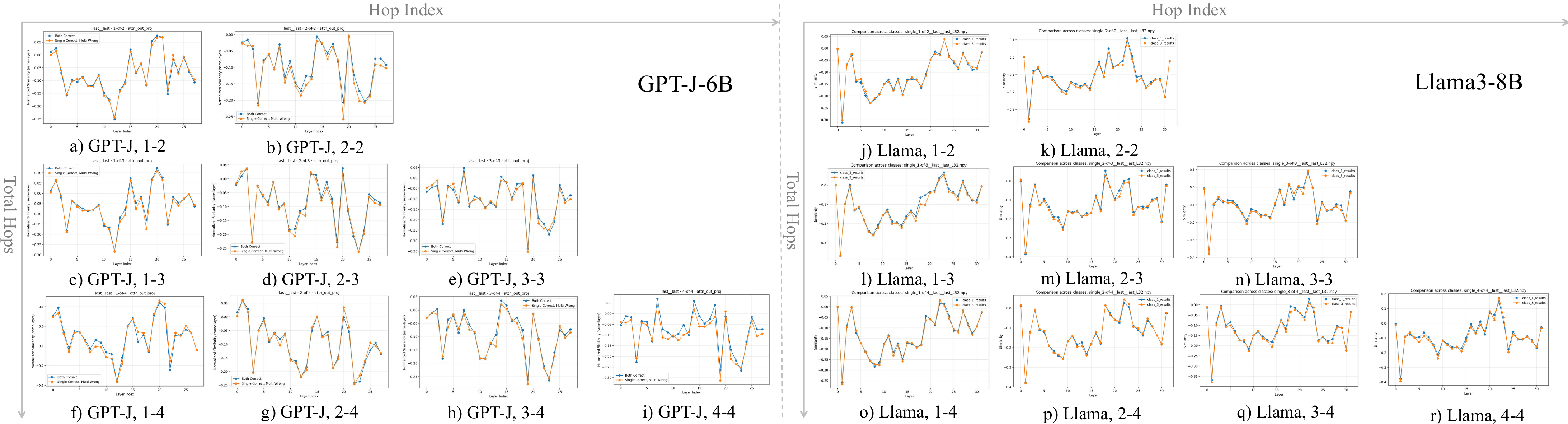}
    \caption{Normalized similarity of \textit{attn\_proj\_out} at the last token between multi-hop queries and their single-hop queries. }
    \label{fig:sim_attn_last_last}
\end{figure*}

\begin{figure*}[!ht]
    \centering
    \includegraphics[width=\textwidth]{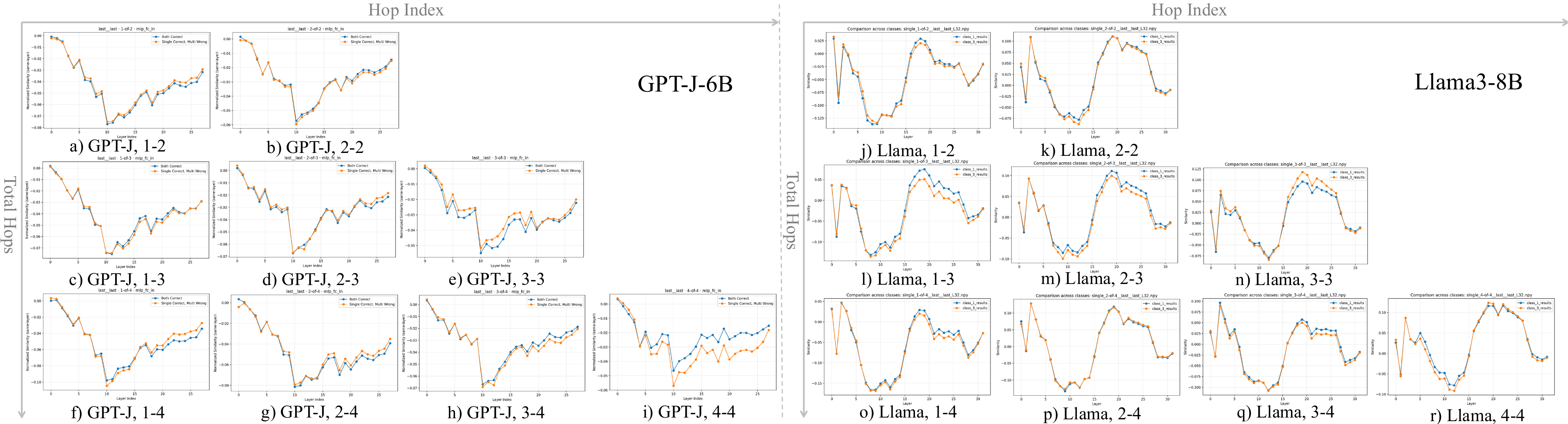}
    \caption{Normalized similarity of \textit{mlp\_fc\_in} at the last token between multi-hop queries and their single-hop queries. }
    \label{fig:sim_mlp_fc_in_last_last}
\end{figure*}

\begin{figure*}[!ht]
    \centering
    \includegraphics[width=\textwidth]{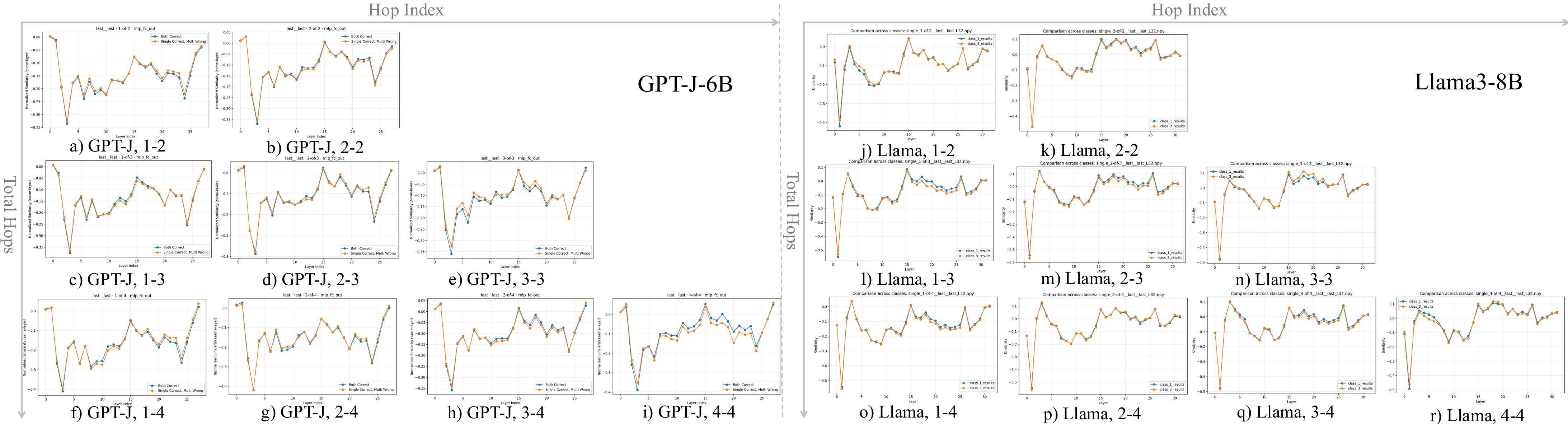}
    \caption{Normalized similarity of \textit{mlp\_fc\_out} at the last token between multi-hop queries and their single-hop queries. }
    \label{fig:sim_mlp_fc_out_last_last}
\end{figure*}

Due to space constraints, we report additional hidden state similarity analyses under different settings in this section. 
At the subject token, we observe consistent patterns across different components in Figure~\ref{fig:sim_attn_sub_sub}-\ref{fig:sim_mlp_fc_out_sub_sub}, aligning with the results obtained from \textit{mlp\_fc\_in} in Figure~\ref{fig:sim_mlp_fc_in_sub_sub}. 
At the final token in Figure~\ref{fig:sim_attn_last_last}-\ref{fig:sim_mlp_fc_out_last_last}, we also observe the same trend discussed in Section~\ref{Sec: Reinterpreting Errors in Multi-hop Reasoning}, the similarity gap between incorrect and correct cases increases as the hop index grows. 
These results further suggest that although the model can correctly answer the corresponding single-hop questions, it fails to sufficiently recall the relevant multi-hop knowledge when answering the full multi-hop query.

\subsection{Average Earliest Layer}
\label{app: Average Earliest Layer}

In Section~\ref{Do LLMs Rely on Explicit Multi-hop Reasoning Circuits?}, we present results under the raw setting, while the results for the filtered settings are included in Table~\ref{tab:earliest_layer_gf} and Table~\ref{tab:earliest_layer_lf} due to space constraints. Consistent with the main results, \emph{layer-order inversion} is observed across all settings, further suggesting that latent multi-hop reasoning in LLMs may not follow explicit reasoning circuits.

\begin{table}[t]
\centering
\resizebox{\linewidth}{!}{
\begin{tabular}{l r r r}
\toprule
\textbf{Hop} & \textbf{Valid} & \textbf{Strict (\%, 95\% CI)} & \textbf{Non-later (\%, 95\% CI)} \\
\midrule
2-hop & 258 & 37.2 [31.5, 43.3] & 42.6 [36.8, 48.7] \\
3-hop & 185 & 46.0 [38.9, 53.1] & 58.9 [51.7, 65.8] \\
4-hop & 116 & 69.8 [61.0, 77.4] & 80.2 [72.0, 86.4] \\
\bottomrule
\end{tabular}
}
\caption{
Instance-level statistics of layer-order inversion on Llama3-8B across different hop counts.
}
\label{tab:instance_inversion}
\end{table}

\subsection{Instance-level Statistics of Layer-Order Inversion}
\label{app:instance_inversion}

To complement the average earliest-layer results in Section~\ref{Do LLMs Rely on Explicit Multi-hop Reasoning Circuits?}, we further examine \emph{layer-order inversion} at the instance level on Llama~3. We consider valid queries where both the first bridge entity $e_1$ and the final-hop entity are decodable, and report the proportions of strict inversion (the final-hop entity is decoded earlier than $e_1$) and non-later inversion (the final-hop entity is decoded no later than $e_1$), together with Wilson 95\% confidence intervals, in Table~\ref{tab:instance_inversion}.

The results show that \emph{layer-order inversion} occurs in a substantial fraction of instances rather than being driven by a few extreme cases. Moreover, both inversion rates increase with the number of hops, indicating that the \emph{layer-order inversion} observed in Section~\ref{Do LLMs Rely on Explicit Multi-hop Reasoning Circuits?} reflects a systematic instance-level tendency and becomes more prevalent in longer reasoning chains.

\subsection{Patchscopes results on Incorrect cases}
\label{app:incorrect_patchscopes}

In Section~\ref{Sec: Reinterpreting Errors in Multi-hop Reasoning}, we use Patchscopes to analyze failures in \textbf{Incorrect} cases.
Due to space constraints, we move some of the Patchscopes results to this appendix.
As shown in Figure~\ref{fig:patchscopes_class3}, deeper-hop entities are decoded less frequently in \textbf{Incorrect} cases at both the subject and last-token positions, compared with the \textbf{Correct} subset in Figure~\ref{fig:patchscopes_class1}.
This provides complementary evidence that incorrect multi-hop predictions can arise from insufficient activation of deeper-hop knowledge under the multi-hop context.

\section{Additional Analyses under the Probabilistic Recall-and-Extract Framework}
\label{app:analyses}
Due to space constraints, this appendix further discusses previously observed experimental phenomena under the probabilistic recall-and-extract framework.

\paragraph{Knockout.}
Knockout~\cite{geva2023dissecting-recall-then-extract} refers to the observation that blocking attention from earlier tokens to the last token in shallow layers significantly affects the probability of generating the correct answer, which has been interpreted as evidence that intermediate entities are propagated to the final position. Under our probabilistic recall-and-extract framework, such intervention reduces the probability of recalling deeper-hop entities at the last token, weakening the intuition-like signal for answer selection and thereby altering the answer distribution.

\paragraph{Back-patching.}
Back-patching~\cite{biran2024hopping-too-late} refers to the intervention of patching hidden states from deeper layers back into shallow layers, which has been shown to improve performance on multi-hop questions. This effect has previously been explained by the view that intermediate entities are propagated too late, which prevents the last token from handling subsequent hops.
Under our probabilistic recall-and-extract framework, back-patching can be understood as compensating for insufficient shallow recall: injecting deeper representations into shallow layers increases the probability of recalling deeper-hop entities, strengthening the intuition-like signal for answer selection and thereby facilitating correct answer generation.

% \paragraph{Chain-of-Thought.}
% Chain-of-thought (CoT) prompting~\cite{wei2022chain-of-thought, renze2024benefits-ccot} improves multi-hop reasoning by explicitly generating intermediate steps, which is difficult to explain under circuit-based views that focus mainly on vertical, layer-wise computation.
% Under our probabilistic recall-and-extract framework, the newly generated tokens (e.g., bridge entities) enrich the horizontal context, 
% \hl{which increases the probability of recalling answer-relevant knowledge in shallow MLP layers, including deeper-hop knowledge that may serve as an intuition-like signal, and shifts the model's output distribution toward the correct answer.}

\paragraph{Chain-of-Thought.}
Chain-of-thought (CoT) prompting~\cite{wei2022chain-of-thought, renze2024benefits-ccot} improves multi-hop reasoning by explicitly generating intermediate steps, which is difficult to explain under circuit-based views that focus mainly on vertical, layer-wise computation.
Under our probabilistic recall-and-extract framework, the newly generated tokens (e.g., bridge entities) strengthen the horizontal recall described in Section~\ref{Mechanism} by enriching the context available to subsequent tokens.
This makes answer-relevant deeper-hop knowledge more likely to be recalled and selected, thereby shifting the model's output distribution toward the correct answer.

\section{Analysis of Missing Cases}
\label{app: missing}

\begin{figure*}[!ht]
    \centering
    \includegraphics[width=\textwidth]{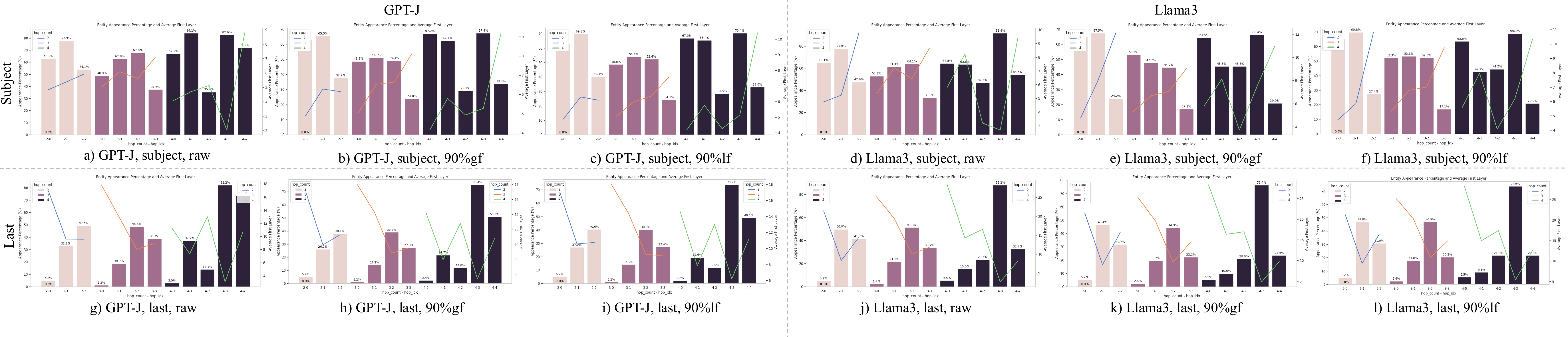}
    \caption{Patchscopes results for GPT-J (left) and Llama3 (right) on the Missing subset.}
    \label{fig:patchscopes_class4}
\end{figure*}

\begin{figure*}[!ht]
    \centering
    \includegraphics[width=\textwidth]{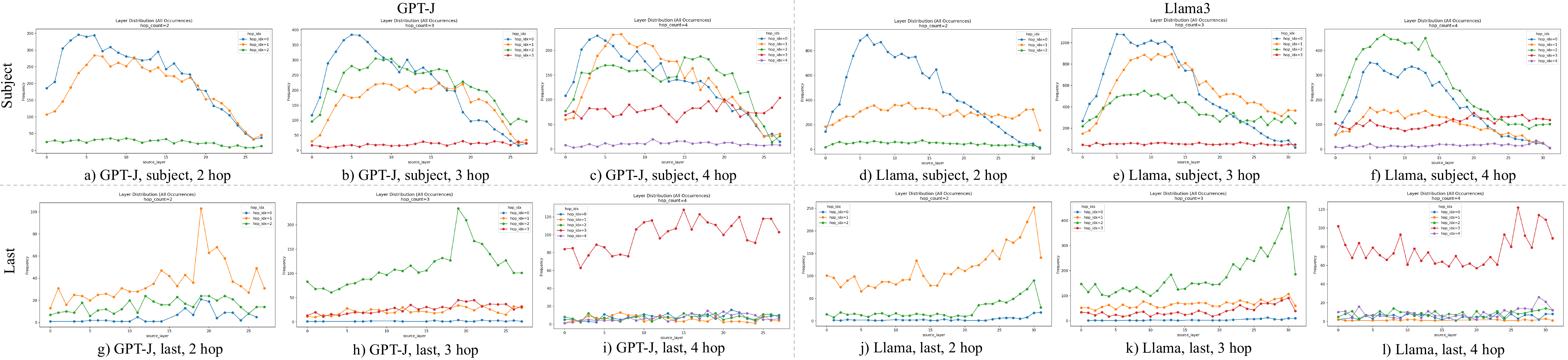}
    \caption{Layer-wise generation distributions produced by Patchscopes on the Missing subset. }
    \label{fig:Patchscopes distributions class4}
\end{figure*}

Due to space constraints, we provide the analysis of the \textbf{Missing} cases in this section. As shown in Figure~\ref{fig:patchscopes_class4} and Figure~\ref{fig:Patchscopes distributions class4}, the decoding rates of early-hop knowledge at the subject position are substantially lower in Missing cases than in both \textbf{Correct} and \textbf{Incorrect} cases. As a result, the model fails to recall the foundational knowledge required for multi-hop reasoning, preventing it from producing the correct answer.

\clearpage

\end{document}